%% file: DR_formulation_arXiv.tex
\documentclass[11pt]{article}

\usepackage{arxiv} 

\usepackage[round]{natbib}
\renewcommand{\bibname}{References}
\renewcommand{\bibsection}{\subsubsection*{\bibname}}

\input{math_commands.tex}  

\usepackage{amsmath}
\usepackage[dvipsnames]{xcolor} %

\usepackage{hyperref}
\usepackage{url}
\usepackage{lineno}

\usepackage{booktabs}       %

\usepackage{pgfplots, amsthm, graphicx, subcaption, dsfont, tikz, tikz-qtree, tabularx, amssymb,enumerate, verbatim, enumitem,makecell,multirow,bm}%
\usepackage[colorinlistoftodos, textwidth=30mm,color=blue!10]{todonotes}%

\usepackage{float} 
\usepackage[bottom]{footmisc} %
\usepackage{footnote}
\usepackage{pdfpages}
\usepackage[noabbrev, capitalise]{cleveref} %

\pgfplotsset{compat=1.11} %
\usepgfplotslibrary{fillbetween}
\usepgfplotslibrary{groupplots}%

\usepackage{authblk}

\newtheorem*{my_pro_setting}{Problem Setting}

\title{Distance-Ratio-Based Formulation for Metric Learning}

\author[1,2]{Hyeongji Kim\thanks{kim.hyeongji@hi.no}}
\author[2]{Pekka Parviainen}
\author[1,2]{Ketil Malde}
\affil[1]{Institute of Marine Research, Bergen, Norway}
\affil[2]{Department of Informatics, University of Bergen, Norway}

\begin{document}
\maketitle

\begin{abstract}
In metric learning, the goal is %
to learn an embedding so that data points with the same class are close to each other and data points with different classes are far apart. %
We propose a distance-ratio-based (DR) formulation for metric learning. %
Like softmax-based formulation for metric learning, it models \(p(y=c|x')\), which is a probability that a query point \(x'\) belongs to a class \(c\). %
The DR formulation has two useful properties. %
First, the corresponding loss %
is not affected by scale changes of an embedding. Second, it outputs the optimal (maximum or minimum) %
classification confidence scores %
on representing points for classes. %
To demonstrate the effectiveness of our formulation, %
we conduct few-shot classification experiments using softmax-based and DR formulations on CUB and \emph{mini}-ImageNet datasets. The results show that DR formulation generally enables faster and more stable metric learning than the softmax-based formulation. As a result, using DR formulation achieves improved or comparable generalization performances. %
\end{abstract}

\section{Introduction}

Modeling %
probability \(p(y=c|x')\), which is a probability that a query point \(x'\) belongs to a class \(c\), plays an important role in discriminative models. Standard neural network based classifiers use the softmax activation function to estimate this probability. %
When \(l_c (x')\) is the logit (pre-softmax) value from the network for class \(c\) and the point \(x'\) and \(\mathcal{Y}\) is a set of classes, the softmax function models \(p(y=c|x')\) as: %
\begin{flalign}\label{eq:std_softmax}
\hat{p}(y=c|x')=\frac{\exp(l_c (x'))}{\sum\limits_{y\in \mathcal{Y}}{\exp(l_y (x'))}},
\end{flalign}
where \(\hat{p}(y=c|x')\) is an estimation of the probability \(p(y=c|x')\).

Standard classifiers work well on classifying %
classes with enough training examples. However, we often encounter %
few-shot classification tasks that we need to classify points from unseen classes with only a few available examples per class. In such cases, standard classifiers may not perform %
well \citep{vinyals2016matching}. %
Moreover, standard classifiers do not %
model similarity %
between different data points on the logit layer.
Metric learning methods %
learn pseudo metrics such that points with the same classes are close, and points with different classes are far apart on the learned embedding spaces.
As a result, 
metric learning models can work well on classifying classes with a few examples \citep{chen2019closer}, and they can be used to find similar data points for each query point \citep{musgrave2020metric}. 

Several metric learning models \citep{goldberger2004neighbourhood, snell2017prototypical, allen2019infinite} use softmax-based formulation to model \(p(y=c|x')\) by replacing logits \(l_c (x')\) in Equation (\ref{eq:std_softmax}) with %
negative squared distances between data points on embedding spaces. %
We found that \emph{1) softmax-based models %
can be affected by scaling embedding space} and thus possibly weaken the training process. %
Moreover, \emph{2) they %
do not have the maximum (or minimum) confidence scores\footnote{Confidence score (value) is an estimated probability of \(p(y=c|x')\) using a model.} on representing points of classes.} %
It implies that when the softmax-based formulation is used for metric learning, %
query points do not directly converge (approach) to points representing the same class on embedding space, and query points do not directly diverge (be far apart) from points representing different classes. As a result, metric learning with %
softmax-based formulation can be unstable.%

To overcome these limitations, we propose an alternative formulation named \emph{distance-ratio-based (DR) formulation} to estimate \(p(y=c|x')\) in metric learning models. Unlike softmax-based formulation, \emph{1) DR formulation is not affected by scaling embedding space.} %
Moreover, \emph{2) it has the maximum confidence score \(1\) on the points representing the same class with query points and the minimum confidence score \(0\) %
on the points representing the different classes.} Hence, when we use DR formulation for metric learning, query points can directly approach to corresponding points and directly diverge from points that represent different classes. 
We analyzed the metric learning %
process with both formulations %
on few-shot learning tasks. Our experimental results show that using our formulation is less likely to be affected by scale changes and more stable. %
As a result, our formulation enables faster training (when Conv4 backbone was used) or comparable training speed (when ResNet18 backbone was used).

\subsection{Problem Settings}

\begin{my_pro_setting}\label{setting1}
Let \(\mathcal{X} \subset \mathbb{R}^{d_{I}}\) be %
an input space, \(\mathcal{Z}=\mathbb{R}^{d_{F}}\) be an unnormalized embedding space, and \(\mathcal{Y}\) be a set of possible classes. The set \(\mathcal{Y}\) also includes classes that are unseen during training. From a joint distribution \(\mathcal{D}\), data %
points \(x\in\mathcal{X}\) and corresponding classes \(c\in\mathcal{Y}\) %
are sampled. %
An embedding function %
\(f_{\theta}: \mathcal{X}\rightarrow \mathcal{Z}\) %
extracts features (embedding vectors) from inputs where \(\theta\) represents learnable parameters. %
We consider the Euclidean distance \(d(\cdot,\cdot)\) on the embedding space \(\mathcal{Z}\). %
\end{my_pro_setting}

In this paper, we only cover unnormalized embedding space \(\mathbb{R}^{d_{F}}\). %
One might be interested in using normalized embedding space \(\mathbb{S}^{\left(d_{F}-1\right)}=\left\{ z\in \mathbb{R}^{d_{F}}| \left\|z\right\|=1 \right\}\). %
For normalized embedding space,  %
one can still use Euclidean distance or angular distance (arc length) as both are proper distances. %

To compare softmax-based and distance-ratio-based formulation that estimate \(p(y=c|x')\) for metric learning, in this work, we use prototypical network  \citep{snell2017prototypical} for explanation and experiments. %
We do this because the prototypical network is one of the simplest metric learning models. 

\subsection{Prototypical Network}

Prototypical network \citep{snell2017prototypical} was devised to solve few-shot classification problems, which require to recognize unseen classes during training %
based on only a few labeled points (support points). It is learned by episode training \citep{vinyals2016matching} whose training batch (called an episode) 
consists of a set of support points and a set of query points.
Support points act as guidelines that represent classes. Query points act as evaluations of a model %
to update the model (embedding function \(f_{\theta}\)) %
in a training phase and to measure few-shot classification performances in a testing phase. %

Using embedding vectors from %
support points, prototypical network calculates a \emph{prototype} \(\mathbf{p}_{c}\) to represent a class \(c\). 
A prototype \(\mathbf{p}_{c}\) is defined as: %
\begin{flalign}
\mathbf{p}_{c}=\frac{1}{|S_{c}|}\sum\limits_{(x_i,y_i)\in S_c}{f_{\theta}(x_i)}, \nonumber
\end{flalign}
 where \(S_{c}\) is a set of support points with class \(c\). (When \(|S_{c}|\) is fixed with \(K=|S_{c}|\), a few-shot learning task %
 is called a \(K\)-shot learning task.)%

We can use the Euclidean distance with a prototype \(\mathbf{p}_{c}\) on the embedding space to estimate how close a query point \(x'\) is %
to a class \(c\). We denote the distance as \(d_{x',c}\). Mathematically, \(d_{x',c}\) is:
\begin{flalign}\label{eq:dist_proto}
d_{x',c}=d(f_{\theta}(x'),\mathbf{p}_{c})
\end{flalign}

Using this distance \(d_{x',c}\), prototypical network estimates the probability \(p(y=c|x')\). %
We will explain later about the softmax-based formulation %
and our formulation for this estimation. Based on the estimated probability, training loss \(L\) is defined as %
the average classification loss (cross-entropy) of query points. The loss \(L\) can be written as:
\begin{flalign}\label{eq:L_proto}
L=-\frac{1}{|Q|}\sum_{(x',y')\in Q}{\log(\hat{p}(y=c|x'))},
\end{flalign} 
where \(Q\) is a set of query points in an episode and \(\hat{p}(y=c|x')\) is an estimation of the probability \(p(y=c|x')\).

Based on the training loss \(L\), we can update %
the embedding function \(f_{\theta}\).

\subsection{%
Metric Learning with Softmax-Based Formulation
}

In the original prototypical network \citep{snell2017prototypical}, the softmax-based formulation was used to model the probability \(p(y=c|x')\). The softmax-based formulation is defined by the softmax (in Equation (\ref{eq:std_softmax})) over negative squared distance \(-d_{x',c}^2\). Thus, the formulation can be written as: 
\begin{flalign}\label{eq:dist_softmax}
\hat{p}(y=c|x')=\frac{\exp(-d_{x',c}^2)}{\sum\limits_{y\in {\mathcal{Y}}_E}{\exp(-d_{x',y}^2)}},
\end{flalign}
where \(\mathcal{Y}_E\) is a subset of \(\mathcal{Y}\) that represents a set of possible classes within an episode. (When \(|{\mathcal{Y}}_E|\) is fixed with \(N=|{\mathcal{Y}}_E|\), a few-shot learning task is called a \(N\)-way learning task.)\\
We denote the value in Equation (\ref{eq:dist_softmax}) as \(\sigma_c (x')\). When we use \(\sigma_c (x')\) to estimate the probability \(p(y=c|x')\), we denote the corresponding loss (defined in Equation (\ref{eq:L_proto})) as \(L_{S}\). 

\newcolumntype{M}[1]{>{\arraybackslash}m{#1}}
\newcolumntype{\Mc}[1]{>{\centering\arraybackslash}m{#1}}

The softmax-based formulation can be obtained by estimating a class-conditional distribution \(p(x'|y=c)\) with a Gaussian distribution. Based on this, in Appendix \ref{sec:represent_mean}, we explain %
why an average point is an appropriate point to represent a class when we use the softmax-based formulation.

\subsubsection{Analysis of Softmax-Based Formulation}

To analyze the formulation in Equation (\ref{eq:dist_softmax}), %
let us consider a toy example. %
In this example, there are only two classes \(c_1\) and \(c_2\) and corresponding prototypes \(\mathbf{p}_{c_1}\) and \(\mathbf{p}_{c_2}\). %
Let us consider a query point \(x'\) that has distance %
\(d_{x',c_1}\) and %
\(d_{x',c_2}\) as in two cases in Table \ref{tab:toy_ex_cases}. When we compare case (a) and 2 times scaled case (case (b)), we can check that the loss \(L_{s}\) is much smaller for the scaled case (\(6.1442\times 10^{-6}\)). %
In other words, 
\emph{simply scaling an embedding can change the confidence score %
and thus corresponding training loss}. It implies that 
embedding can be scaled to reduce training loss. Thus, using softmax-based models may weaken a training process by allowing unnecessary model updates that do not change relative locations of data points.

\begin{figure*}[tbp]
    \centering
    \vspace{.05in}
    \begin{subfigure}[b]{0.27\textwidth}
        \includegraphics[width=1.0\linewidth,height=0.75\linewidth]{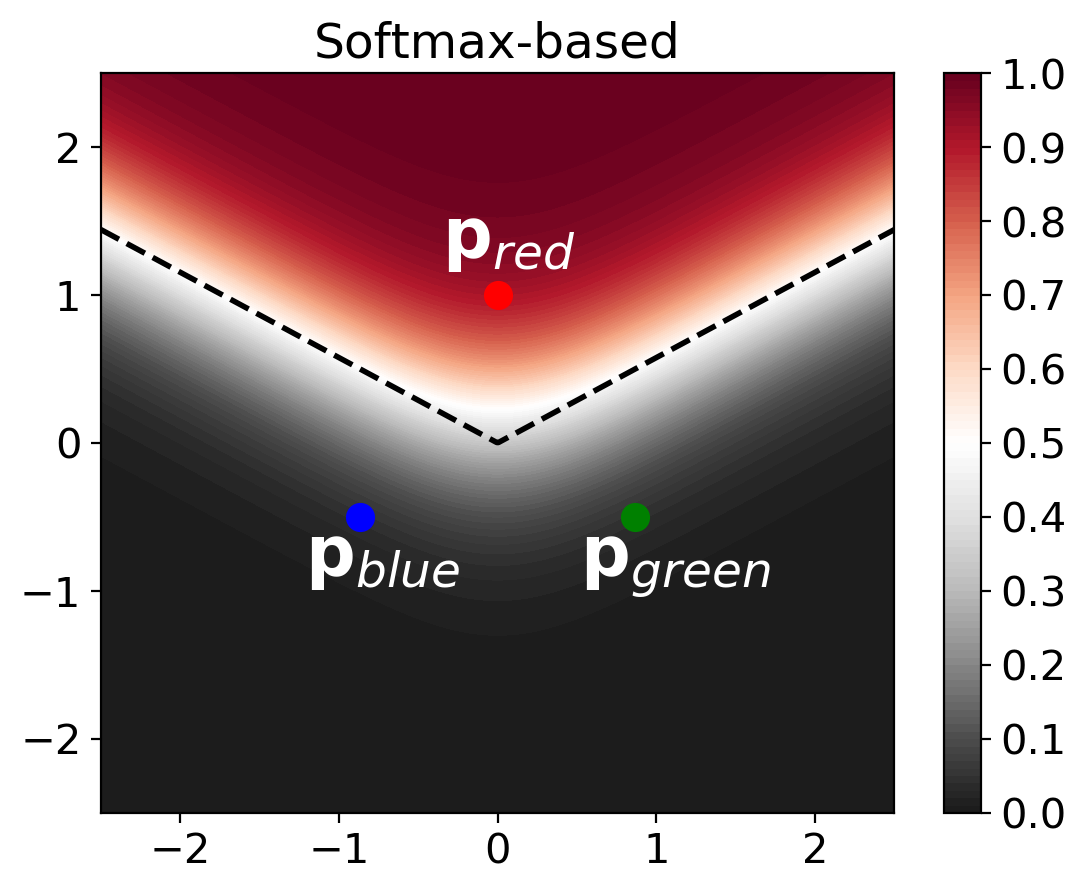}%
        \caption{\(\sigma_{\textcolor{red}{red}} (x')\)
        }\label{fig:p_red_x'_b}%
    \end{subfigure}
    \begin{subfigure}[b]{0.27\textwidth}
        \includegraphics[width=1.0\linewidth,height=0.75\linewidth]{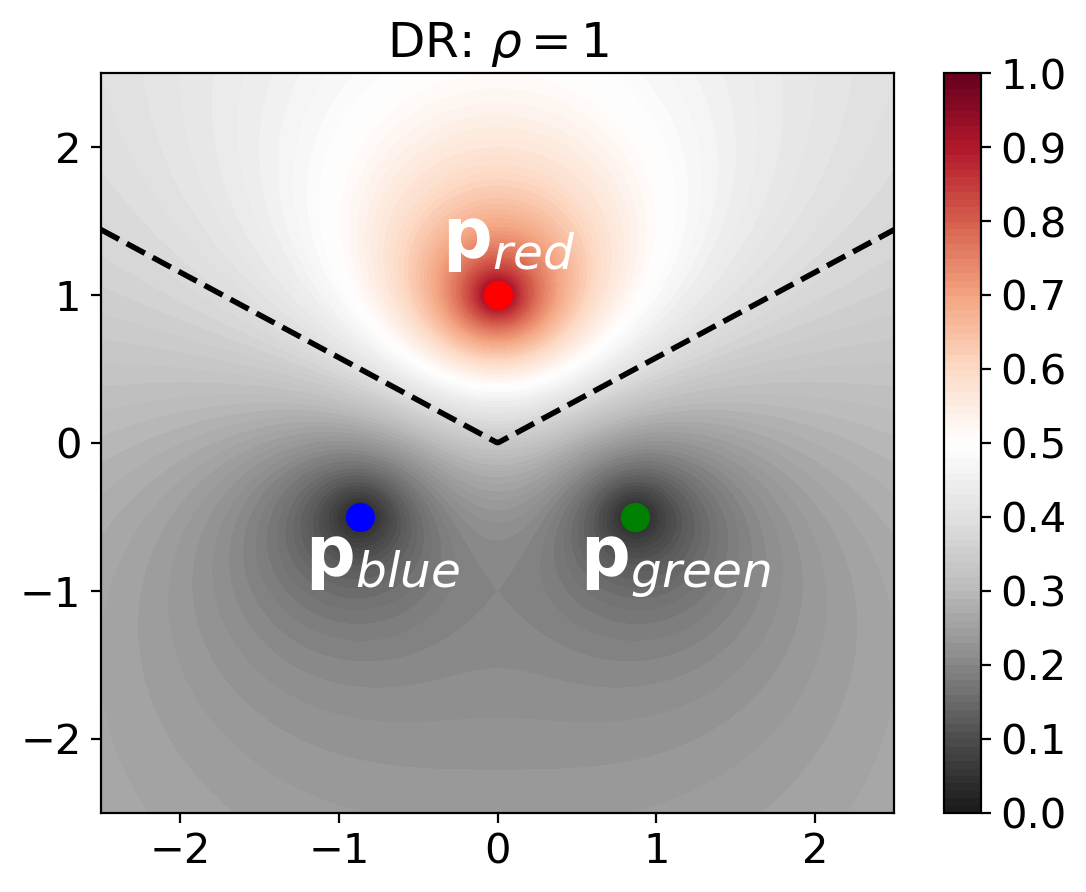}%
        \caption{\(\delta_{\textcolor{red}{red}} (x')\) %
        with \(\rho=1\)}\label{fig:p_red_x'_c}%
    \end{subfigure}
    \begin{subfigure}[b]{0.27\textwidth}
        \includegraphics[width=1.0\linewidth,height=0.75\linewidth]{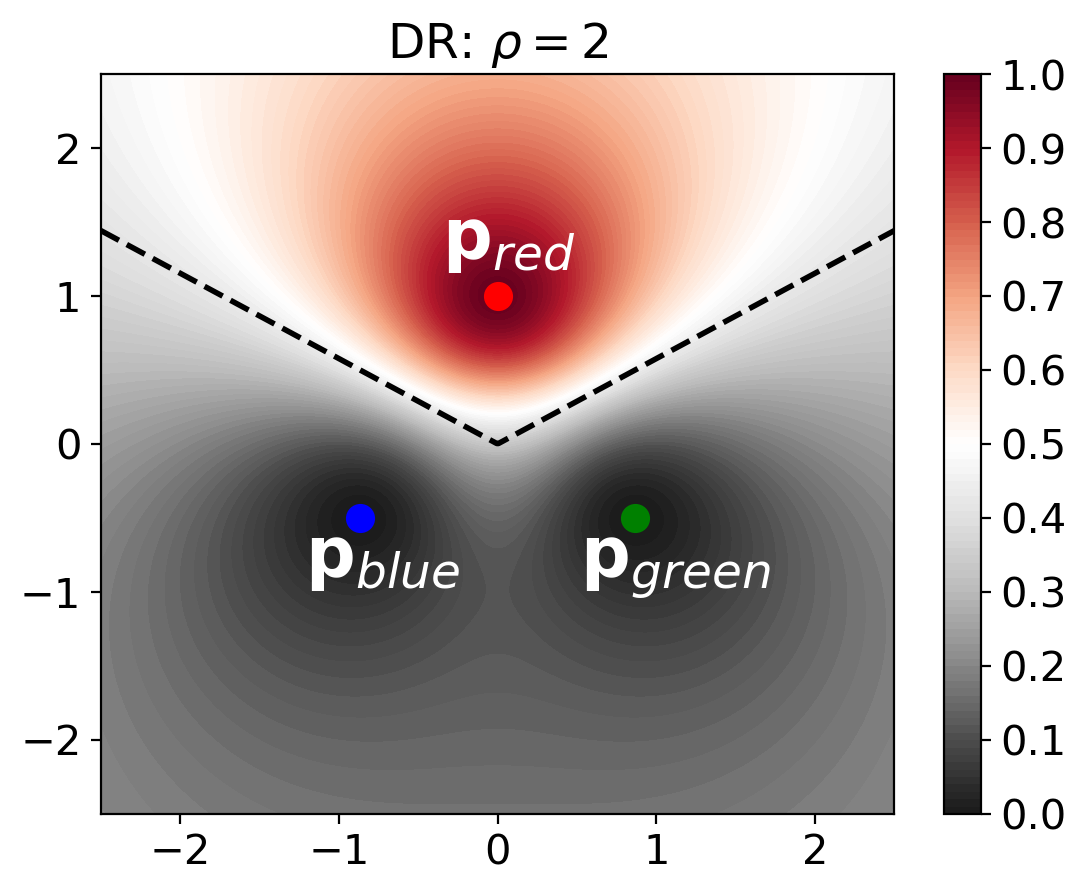}%
        \caption{\(\delta_{\textcolor{red}{red}} (x')\) %
        with \(\rho=2\)}\label{fig:p_red_x'_d}%
    \end{subfigure}
    \caption{Visualization of estimations of \(p(y=\textcolor{red}{red}|x')\) based on softmax-based formulation and distance-ratio-based formulation. %
    }
    \label{fig:p_red_x'}
\end{figure*}

To inspect the locations that maximize or minimize confidence scores, in Figure \ref{fig:p_red_x'}, 
we visualized the estimated probability \(\hat{p}(y=\textcolor{red}{red}|x')\) using three prototypes \(\textcolor{red}{\mathbf{p}_{red}}\), \(\textcolor{Green}{\mathbf{p}_{green}}\), and \(\textcolor{blue}{\mathbf{p}_{blue}}\).
For the %
softmax-based model in Figure \ref{fig:p_red_x'_b} \citep{goldberger2004neighbourhood, snell2017prototypical, allen2019infinite}, %
\emph{the maximum confidence value is not even at the prototype} \(\textcolor{red}{\mathbf{p}_{red}}\). %
It implies %
that when we train an embedding with training loss \(L_S\), query points with the red class do not converge directly to the prototype %
\(\textcolor{red}{\mathbf{p}_{red}}\). %
In Figure %
\ref{fig:p_red_x'_b}, the prototypes %
with different classes \(\textcolor{Green}{\mathbf{p}_{green}}\) %
and \(\textcolor{blue}{\mathbf{p}_{blue}}\) %
are not the points that minimize confidence values. It means that query points with red class do not directly diverge (get far apart) from prototypes \(\textcolor{Green}{\mathbf{p}_{green}}\) %
and \(\textcolor{blue}{\mathbf{p}_{blue}}\). %
As prototypes do not provide direct guidelines for query points, metric learning with softmax-based formulation can be unstable.

\begin{table*}[hbp]%
    \caption{A toy example with different \(d_{x',c_1}\) and \(d_{x',c_2}\), and the corresponding values. We assumed the query point \(x'\) has class \(c_1\) to calculate the losses. In this table, we set \(\rho=2\) for DR formulation. \(\delta\) is defined in Section \ref{sec:DR_form}. }
    \label{tab:toy_ex_cases}
    \centering
    \begin{tabular}{M{1.5cm}|\Mc{1.5cm}\Mc{1.5cm}|\Mc{1.75cm}\Mc{1.75cm}|\Mc{2.cm}\Mc{2.cm}}
    \toprule
    \scriptsize Cases & \scriptsize \(d_{x',c_1}\)& \scriptsize \(d_{x',c_2}\)& \scriptsize \(\sigma_{c_1}(x')\)& \scriptsize \(\delta_{c_1}(x')\)& \scriptsize \(L_{s}\)& \scriptsize \(L_{DR}\) \\ \midrule
    \scriptsize {Case (a)}&\scriptsize {\(1\)} &\scriptsize {\(2 \)} & \scriptsize \( 0.95257\) & \scriptsize \( 0.80000\) & \scriptsize \( 0.048587 \) & \scriptsize \( 0.22314\) \\  \hline
    \scriptsize {Case (b)}&\scriptsize {\(2\)} &\scriptsize \( 4\)
    &\scriptsize \(0.99999\) & \scriptsize \( 0.80000 \)& \scriptsize \( 6.1442\times 10^{-6}
    \) & \scriptsize \( 0.22314 \) \\ %
    \bottomrule
    \end{tabular}
\end{table*}

\section{Metric Learning with Distance-Ratio-based Formulation%
}\label{sec:DR_form}

To handle %
the limitations of softmax-based formulation in metric learning, %
we propose an alternative form called \emph{distance-ratio-based (DR) formulation} for %
estimating probability \(p(y=c|x')\). When we use distance \(d_{x',c}\) as in Equation (\ref{eq:dist_proto}), %
DR formulation is defined as: 
\begin{flalign}\label{eq:DR_form}
\hat{p}(y=c|x')=\frac{ \frac{1}{d_{x',c}^\rho} }{\sum\limits_{y\in {\mathcal{Y}_E}}{ \frac{1}{d_{x',y}^\rho} }}=\frac{ {d_{x',c}}^{-\rho} }{\sum\limits_{y\in {\mathcal{Y}_E}}{ {d_{x',y}}^{-\rho} }},
\end{flalign}
where %
\(\rho>0\) is a learnable parameter. %
When \(d_{x',c}=0\) %
and \(d_{x',c'}>0\) %
for all \(c'\neq c\), we define \(\hat{p}(y=c|x')\) as \(1\) and \(\hat{p}(y=c'|x')\) as \(0\). As this formulation uses ratios of distances for classification, we call it as \emph{distance-ratio-based formulation}. (One can check that Equation (\ref{eq:DR_form}) can be obtained by replacing the negative squared distance \(-d_{x',c}^2\) in Equation (\ref{eq:dist_softmax}) 
with \(-\rho \ln(d_{x',c})\).)

Let us denote the value in Equation (\ref{eq:DR_form}) as \(\delta_c (x')\). %
Then, when we use DR formulation to estimate the probability \(p(y=c|x')\), we denote the corresponding training loss (defined in Equation (\ref{eq:L_proto})) as \(L_{DR}\). 
Based on the training loss \(L_{DR}\), we can update %
the embedding function \(f_{\theta}\) and also the learnable parameter \(\rho\).

\subsection{Analysis of Distance-Ratio-Based Formulation}

To analyze our formulation, let us consider when we change the scale of an embedding space. When we scale embedding with a scale parameter \(\alpha>0\), then the corresponding estimation of probability \(p(y=c|x')\) with DR formulation is: 
\begin{flalign}\label{eq:DR_form_scale}
\hat{p}(y=c|x')=&\frac{ \frac{1}{\left(\alpha d_{x',c}\right)^\rho} }{\sum\limits_{y\in {\mathcal{Y}_E}}{ \frac{1}{\left(\alpha d_{x',y}\right)^\rho} }}=\frac{ \frac{1}{\alpha^\rho d_{x',c}^\rho} }{\sum\limits_{y\in {\mathcal{Y}_E}}{ \frac{1}{\alpha^\rho d_{x',y}^\rho} }}\nonumber\\
=&\frac{ \frac{1}{\alpha^\rho} \frac{1}{d_{x',c}^\rho} }{ \frac{1}{\alpha^\rho}\sum\limits_{y\in {\mathcal{Y}_E}}{  \frac{1}{d_{x',y}^\rho} }}=\frac{ \frac{1}{d_{x',c}^\rho} }{\sum\limits_{y\in {\mathcal{Y}_E}}{ \frac{1}{d_{x',y}^\rho} }}
\end{flalign}

Equation (\ref{eq:DR_form_scale}) shows that when we use our formulation, \emph{scaling an embedding has no effect on the confidence scores %
and thus the training loss}. %
(This property can also be checked from the cases (a) and (b) %
in Table \ref{tab:toy_ex_cases}.) 

In addition to the scale invariance property, 
\(\delta_c (x')\) has an additional property that has \emph{optimal confidence scores %
on prototypes.%
} In detail, 
if we assume \(d(\mathbf{p}_{c},\mathbf{p}_{c'})>0\)  %
for \(\forall c'\in \mathcal{Y}_{E}\) with \(c'\neq c\), then the following two equations hold:
\begin{flalign}
\lim_{x'\rightarrow\mathbf{p}_{c}}{\delta_{c}(x')}=1 \label{eq:sparse_1}\\
\lim_{x'\rightarrow\mathbf{p}_{c'}}{\delta_{c}(x')}=0 \label{eq:sparse_0}
\end{flalign}
This property can be checked from Figures \ref{fig:p_red_x'_c} %
and \ref{fig:p_red_x'_d} %
that visualize the estimated probability \(\hat{p}(y=\textcolor{red}{red}|x')\) using %
DR formulation. %
Proof of the property is in Appendix \ref{sec:sparsity_proof}. %
As prototypes %
provide optimal guidelines for query points, %
when we used DR formulation, query points can easily get close to the prototypes %
with their corresponding classes (\(\mathbf{p}_{c}\)) and get far away from the prototypes %
with different classes (\(\mathbf{p}_{c'}\)). Hence, metric learning with DR formulation can be stable.

\section{Experiments}

\subsection{Experiment Settings}

In our experiments, we wanted to investigate the effectiveness of the distance-ratio-based (DR) formulation compared to the softmax-based formulation. %
For that, we trained prototypical networks \citep{snell2017prototypical} based on two formulations for each experiment: %
softmax-based (ProtoNet\_S) and DR formulation (ProtoNet\_{DR}).

We implement 1-shot and 5-shot learning tasks with five classes for each episode (5-way). %
Details of the settings are described in the following paragraphs. %
Codes for our experiments are available in \url{https://github.com/hjk92g/DR_Formulation_ML}.

\paragraph{Dataset} %
We conduct experiments using two common benchmarks for few-shot classification tasks: CUB (200-2011) \citep{wah2011caltech} and \emph{mini}-ImageNet dataset \citep{vinyals2016matching}. The CUB dataset has 200 classes and 11,788 images. %
We use the same 100 training, 50 validation, and 50 test classes split as \citet{chen2019closer}. %
The \emph{mini}-ImageNet dataset is a subset of the ImageNet dataset \citep{deng2009imagenet} suggested by \citet{vinyals2016matching}. It has 100 classes and 600 images per class. %
We use the same 64 training, 16 validation, and 20 test classes split as %
\citet{ravi2016optimization, chen2019closer}. For both datasets, we %
apply data augmentation for training data. Applied data augmentation includes random crop, left-right flip, and color jitter. 

\paragraph{Backbone (Architecture)}
We use two different backbones as embedding functions \(f_{\theta}\) for each experiment: Conv4 \citep{snell2017prototypical} and ResNet18 \citep{he2016deep}. The Conv4 consists of four convolutional blocks. Each block is composed of a 64-filter \(3\times 3\) convolution, batch normalization, a ReLU activation function, and a \(2\times 2\) max-pooling layer. It takes \(84\times84\) sized color images and outputs 1600 dimensional embedding vectors. %
The ResNet18 backbone is the same as in \citet{he2016deep}. It contains convolutions, batch normalizations, ReLU activation functions like the Conv4, but it also has skip connections. %
It takes \(224\times224\) sized color images and outputs 512 dimensional embedding vectors. 

\begin{figure*}[t]
    \centering
    \begin{subfigure}[b]{0.24\textwidth}
        \includegraphics[width=1.0\linewidth,height=0.625\linewidth]{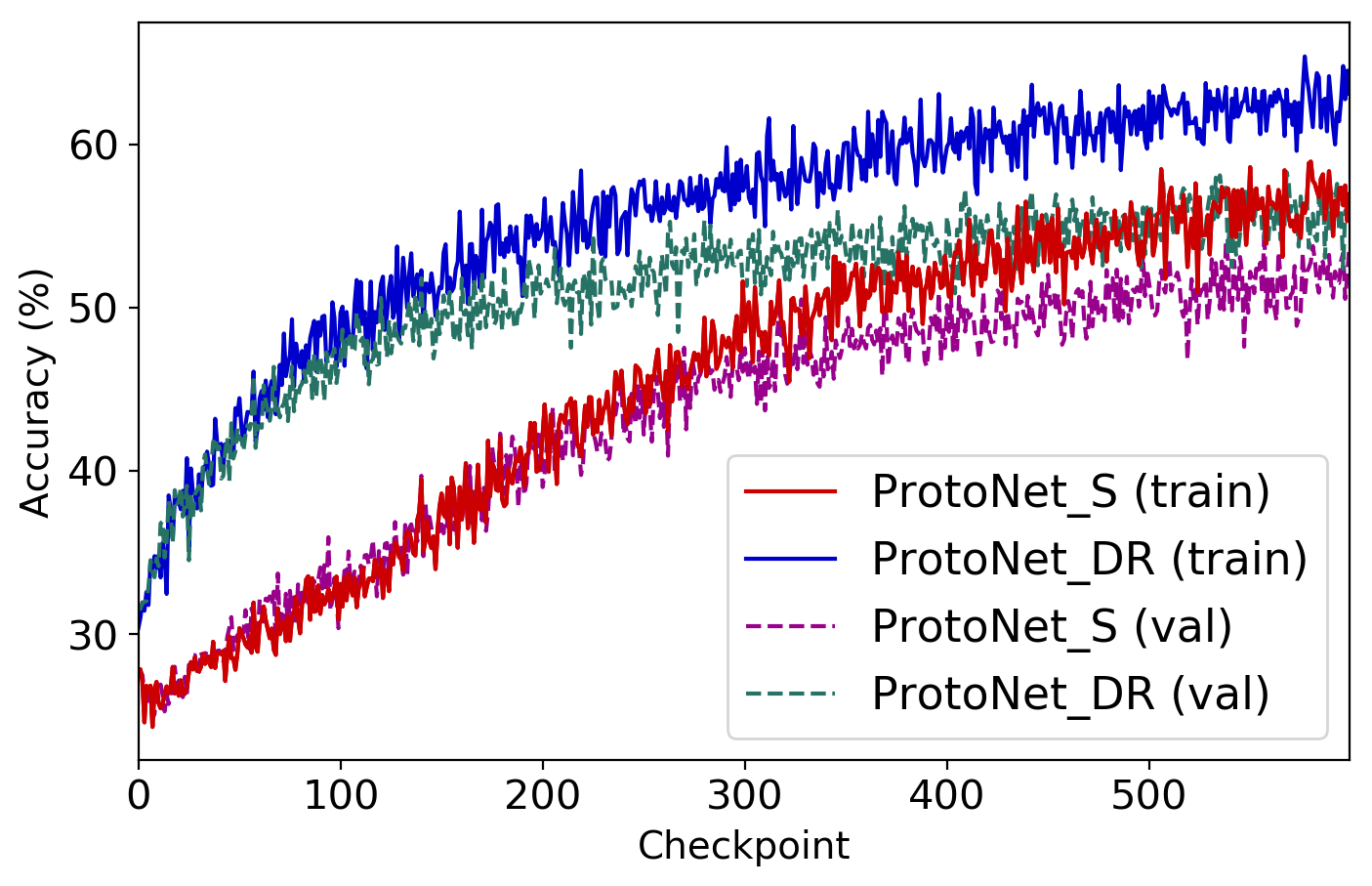} %
        \caption{Data: CUB, Backbone: Conv4}\label{fig:1-shot_training_a}
    \end{subfigure}
    \begin{subfigure}[b]{0.24\textwidth}
        \includegraphics[width=1.0\linewidth,height=0.625\linewidth]{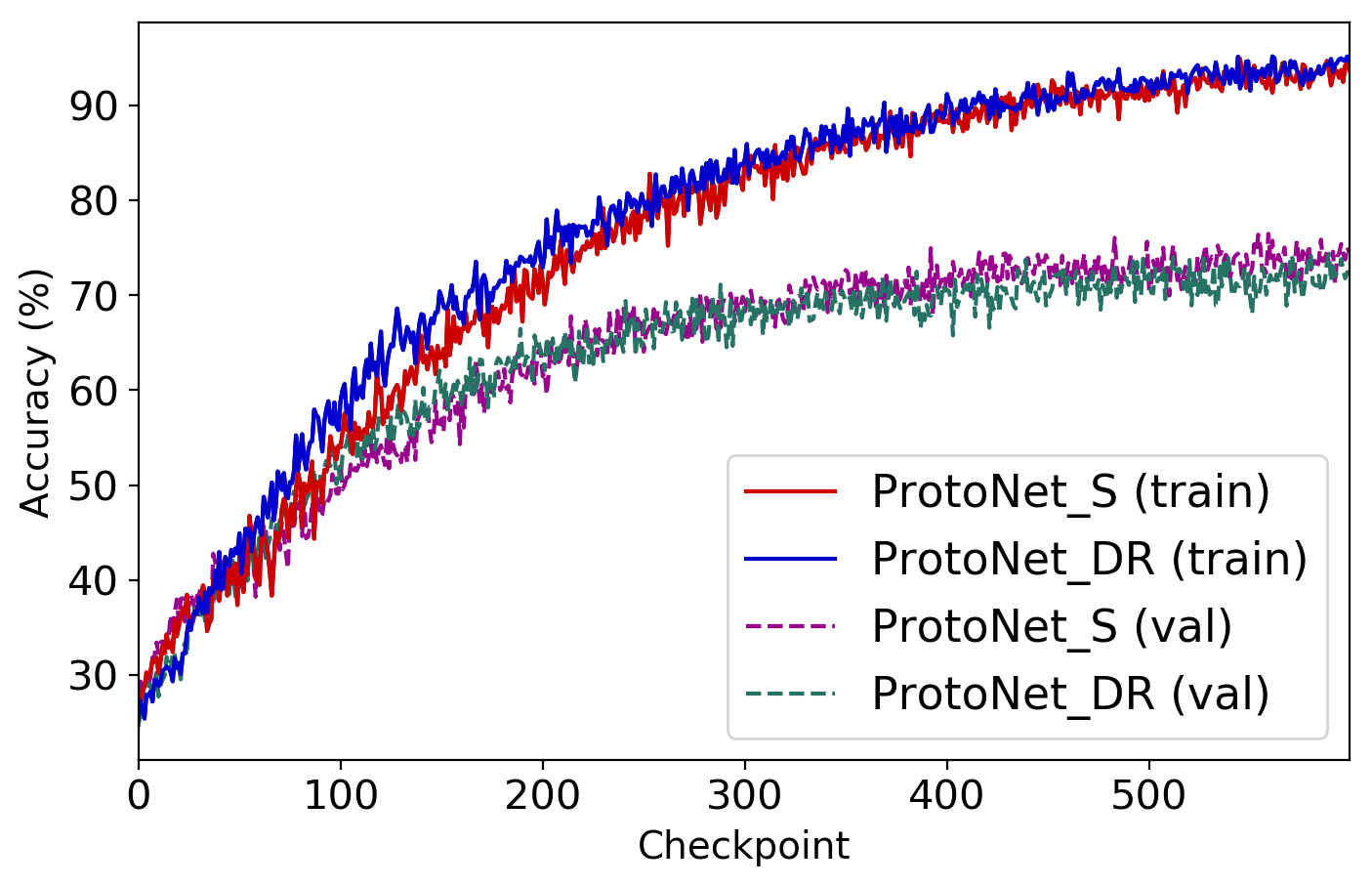}
        \caption{Data: CUB, Backbone: ResNet18}\label{fig:1-shot_training_b}
    \end{subfigure}
    \begin{subfigure}[b]{0.24\textwidth}
        \includegraphics[width=1.0\linewidth,height=0.625\linewidth]{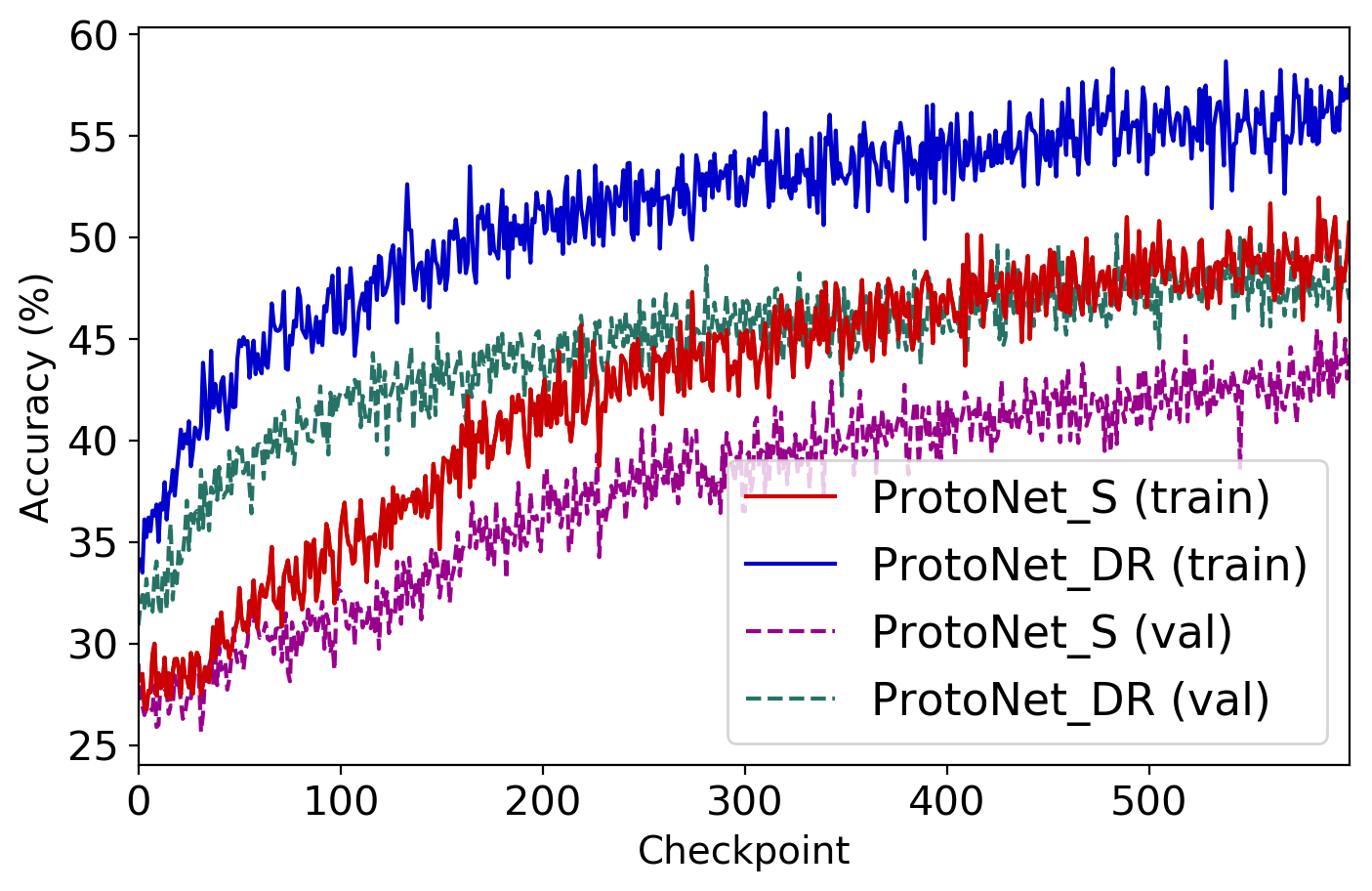} %
        \caption{Data: \emph{mini}-ImageNet, Backbone: Conv4}\label{fig:1-shot_training_c}
    \end{subfigure}
    \begin{subfigure}[b]{0.24\textwidth}
        \includegraphics[width=1.0\linewidth,height=0.625\linewidth]{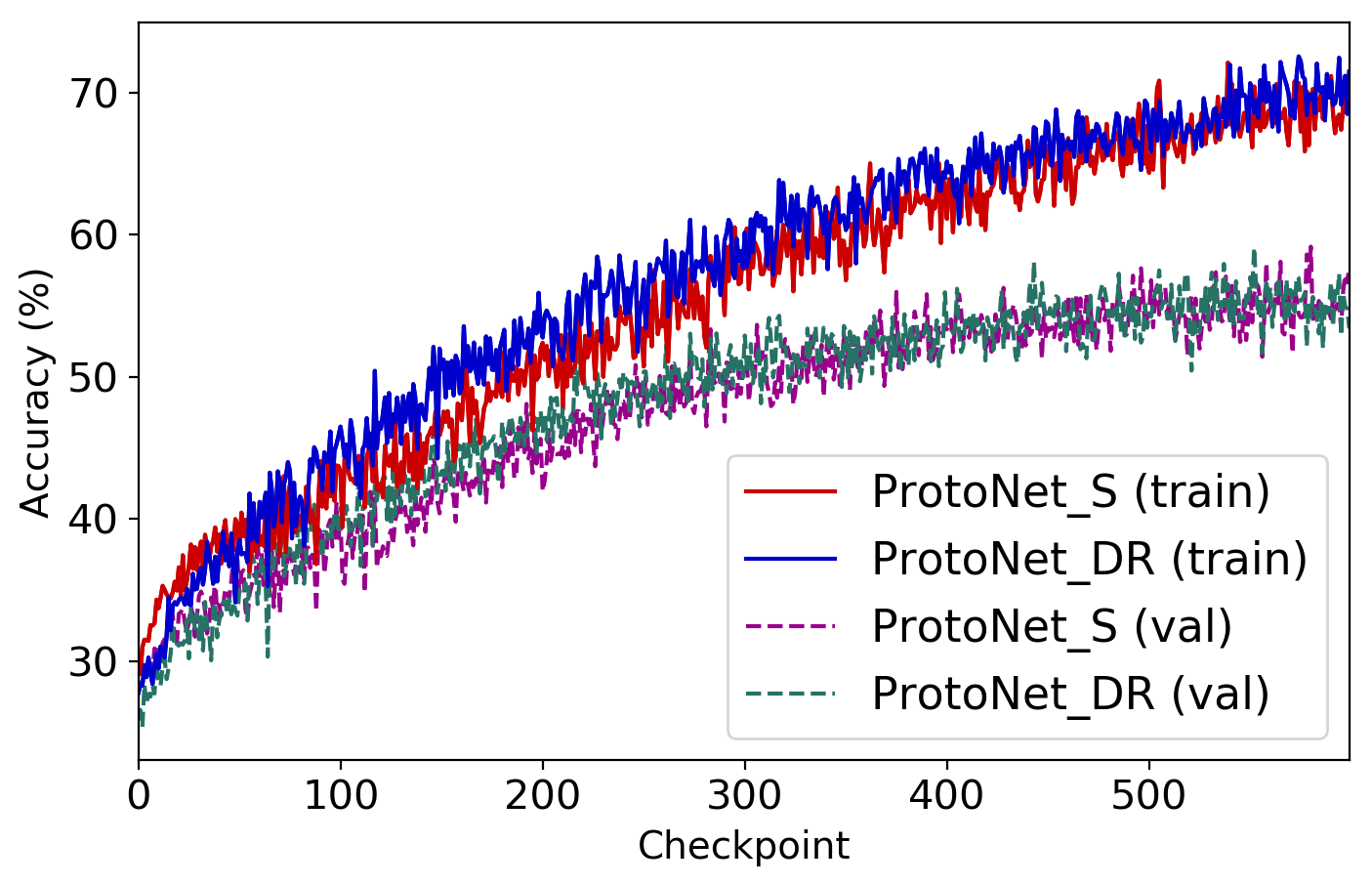}
        \caption{Data: \emph{mini}-ImageNet, Backbone: ResNet18}\label{fig:1-shot_training_d}
    \end{subfigure}
    \caption{Training and validation accuracy curves for two different backbones on 1-shot learning tasks.}
    \label{fig:1-shot_training}
\end{figure*}

\begin{figure*}[t]	
    \centering
    \vspace{.05in}
    \begin{subfigure}[b]{0.24\textwidth}
        \includegraphics[width=1.0\linewidth,height=0.625\linewidth]{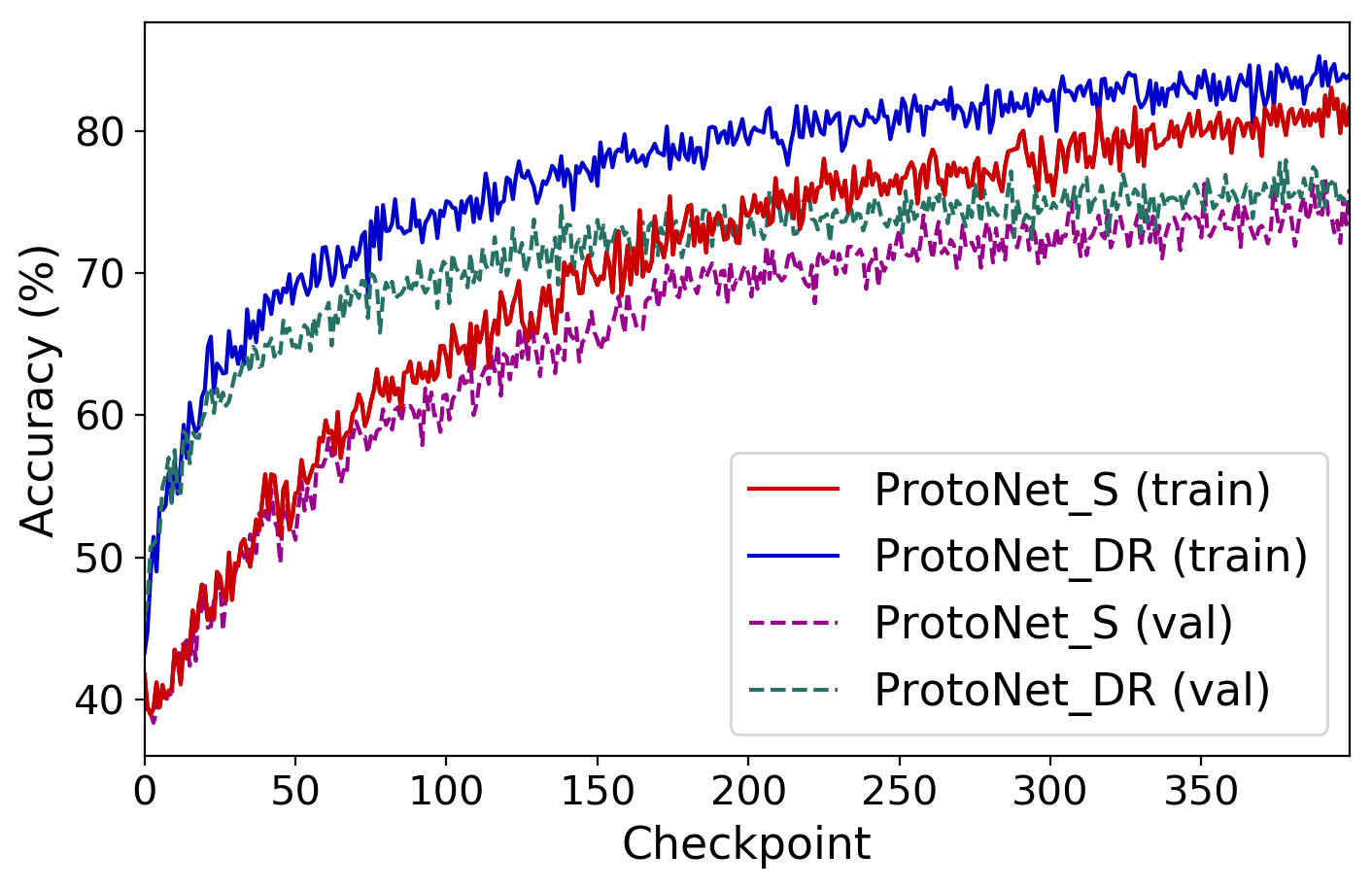} %
        \caption{Data: CUB, Backbone: Conv4}\label{fig:5-shot_training_a}
    \end{subfigure}
    \begin{subfigure}[b]{0.24\textwidth}
        \includegraphics[width=1.0\linewidth,height=0.625\linewidth]{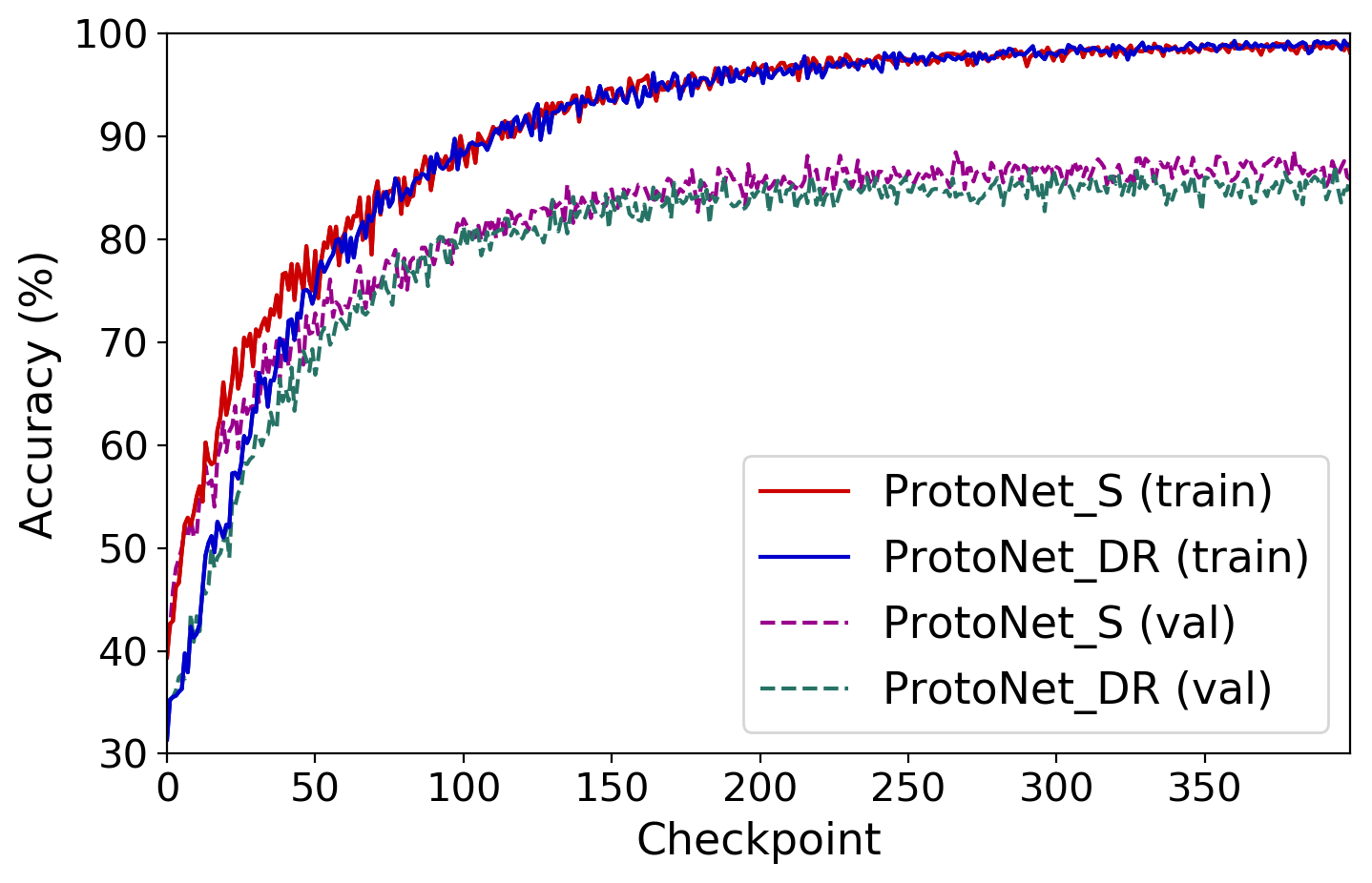}
        \caption{Data: CUB, Backbone: ResNet18}\label{fig:5-shot_training_b}
    \end{subfigure}
    \begin{subfigure}[b]{0.24\textwidth}
        \includegraphics[width=1.0\linewidth,height=0.625\linewidth]{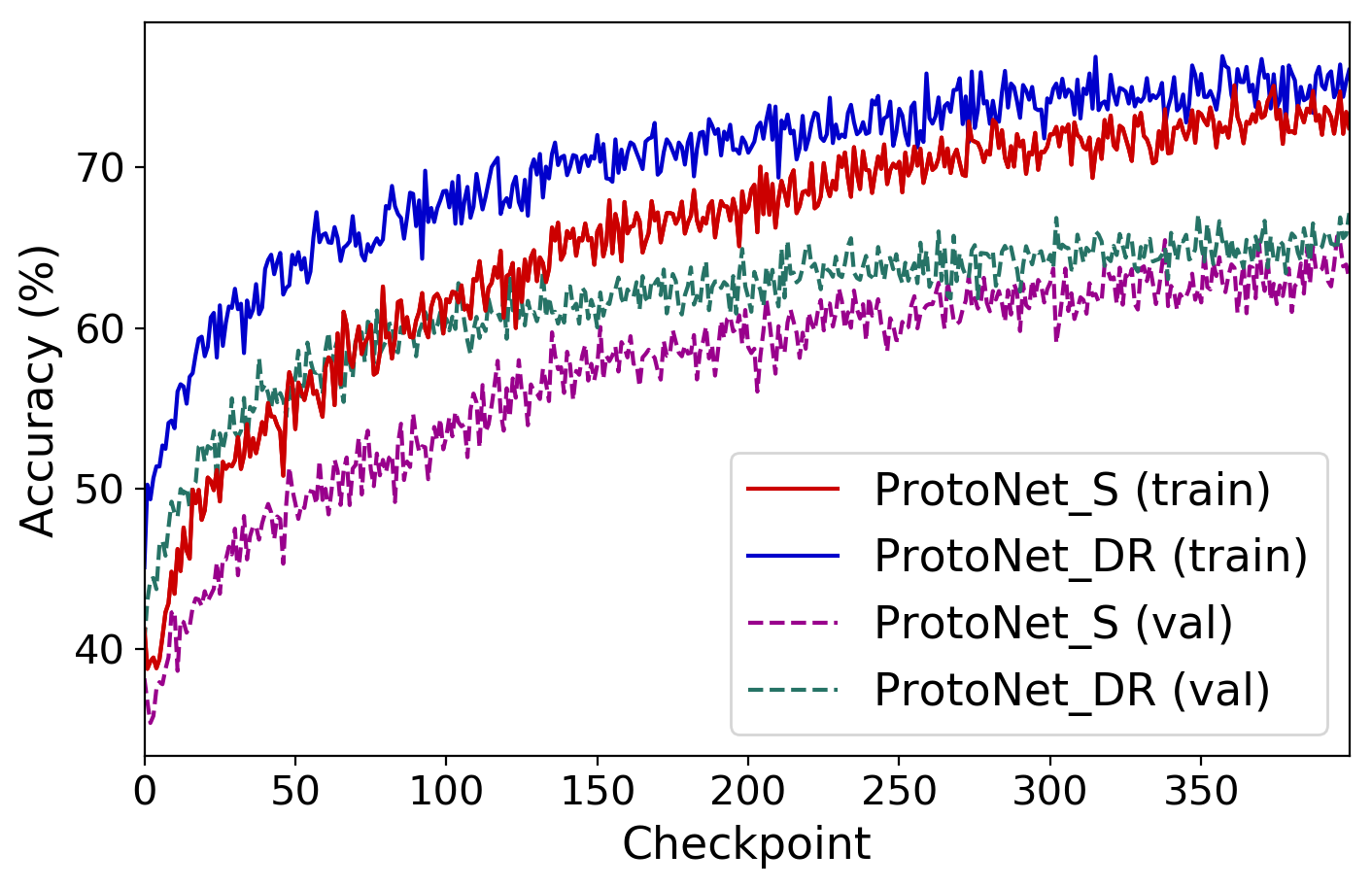} %
        \caption{Data: \emph{mini}-ImageNet, Backbone: Conv4}\label{fig:5-shot_training_c}
    \end{subfigure}
    \begin{subfigure}[b]{0.24\textwidth}
        \includegraphics[width=1.0\linewidth,height=0.625\linewidth]{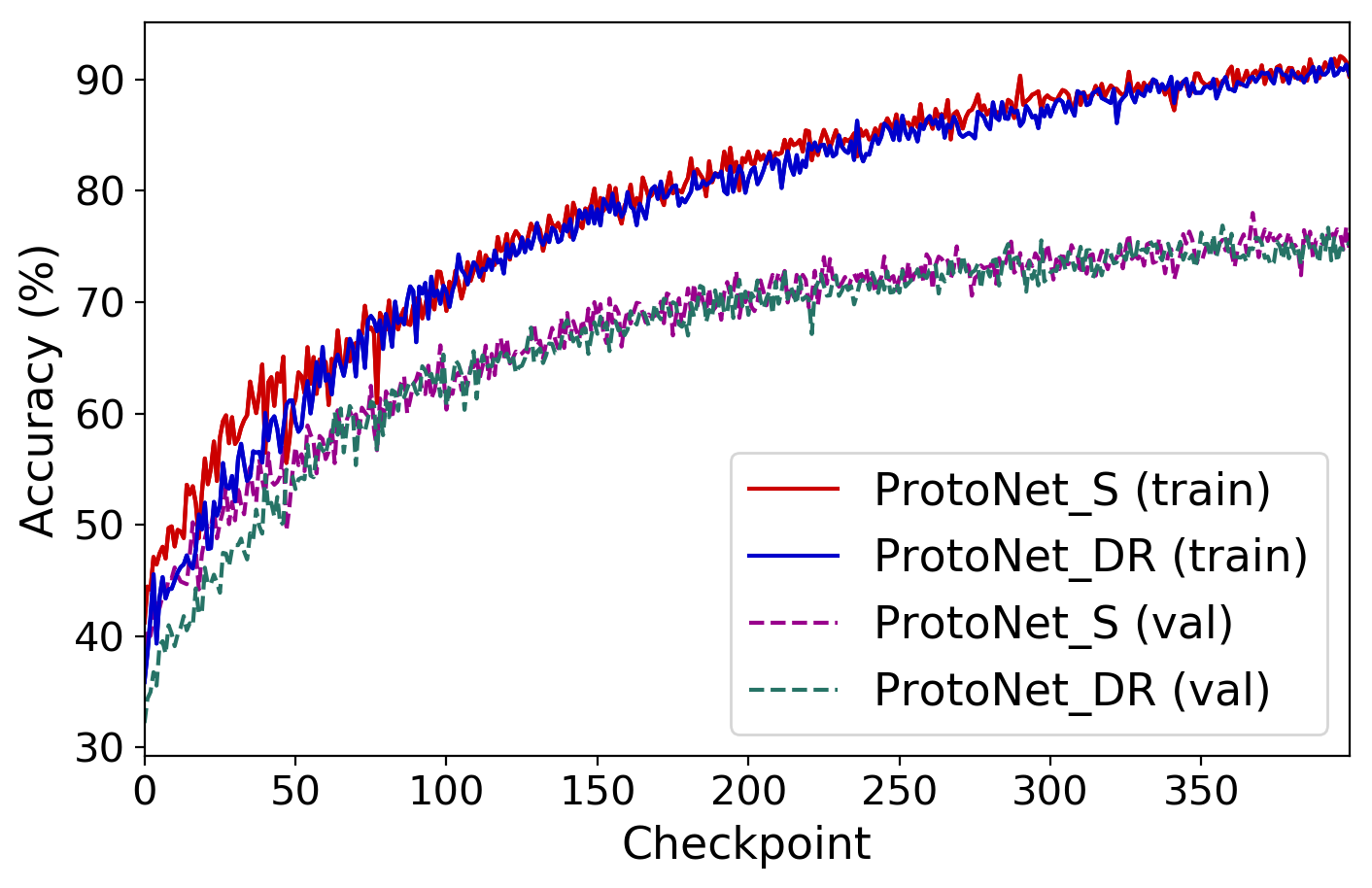}
        \caption{Data: \emph{mini}-ImageNet, Backbone: ResNet18}\label{fig:5-shot_training_d}
    \end{subfigure}
    \caption{Training and validation accuracy curves for two different backbones on 5-shot learning tasks. %
    }
    \label{fig:5-shot_training}
\end{figure*}

\paragraph{Optimization}
Backbones are trained from random weights. %
We use Adam optimizer \citep{kingma2014adam} with a learning rate \(10^{-3}\). To investigate %
training steps, we save training information and validation accuracy for each 100 training episodes, and we call each of these steps a checkpoint. %
For 1-shot classification tasks, we train embedding for 60,000 episodes (600 checkpoints). For 5-shot classification tasks, we train embedding for 50,000 episodes (500 checkpoints). %
Based on validation accuracies on each checkpoint, we select the best model among the %
checkpoints. 

To implement our DR formulation, we modify %
the implementation of the standard softmax-based prototypical network \citep{snell2017prototypical} by replacing negative squared distance \(-d_{x',c}^2\) %
in Equation (\ref{eq:dist_softmax}) by \(-\rho  \ln(d_{x',c})\). %
For numerical stability, we add a small positive value \(10^{-10}\) before taking square root in the calculation of Euclidean distance $d_{x';c}$. %
For the DR formulation, we use \(\ln(\rho)\in\mathbb{R}\) to model \(\rho=\exp(\ln(\rho))\). We set the initial parameter for \(\ln(\rho)\) as \(2.0\). %
Based on this initial value, \(\log(\rho)\) value is trained for all experiments. 

To analyze the local training steps, for each checkpoint (every 100 episodes %
of training), we checked the positions of episode points (both support and query points) on embedding space just before the weight updates and right after the weight updates. When we consider positions on embedding space, we denote a matrix that represents the original positions of episode points as \(X_{origin}\) and the corresponding matrix with updated weights as \(X_{new}\). We assume these matrices are mean-centered. %
When the matrices are not mean-centered, we center the matrices so that an average point is located at zero. %
Then, we model the matrix \(X_{new}\) as a modification of \(\alpha^* X_{origin}\) for an unknown scale parameter \(\alpha^*\). Based on this model on \(X_{new}\), %
we calculated a score called \emph{norm ratio} \(\phi\) that measures the relative effect of scaling (\(0\le \phi\le 1\)). It is defined as:
\begin{flalign}\label{eq:norm_ratio_main}
\phi = \frac{\left \|  X_{new}-\hat{\alpha^*} X_{origin}\right \|_F}{\left \| X_{new}-X_{origin} \right \|_F},
\end{flalign}
where \(\left \| \cdot \right \|_F\) is the Frobenius norm and \(\hat{\alpha^*}\) is an estimated scaling parameter by minimizing the numerator of Equation (\(\ref{eq:norm_ratio_main})\). %
Norm ratio \(\phi\) is close to \(0\) when the major changes are due to scaling. Norm ratio \(\phi\) is close to \(1\) when a magnitude of an embedding is not changed. %
A detailed explanation for the norm ratio is in Appendix \ref{sec:proposed_norm_ratio}. 

\begin{figure*}[t]
    \centering%
    \begin{subfigure}[b]{0.24\textwidth}%
        \includegraphics[width=1.0\linewidth,height=0.625\linewidth]{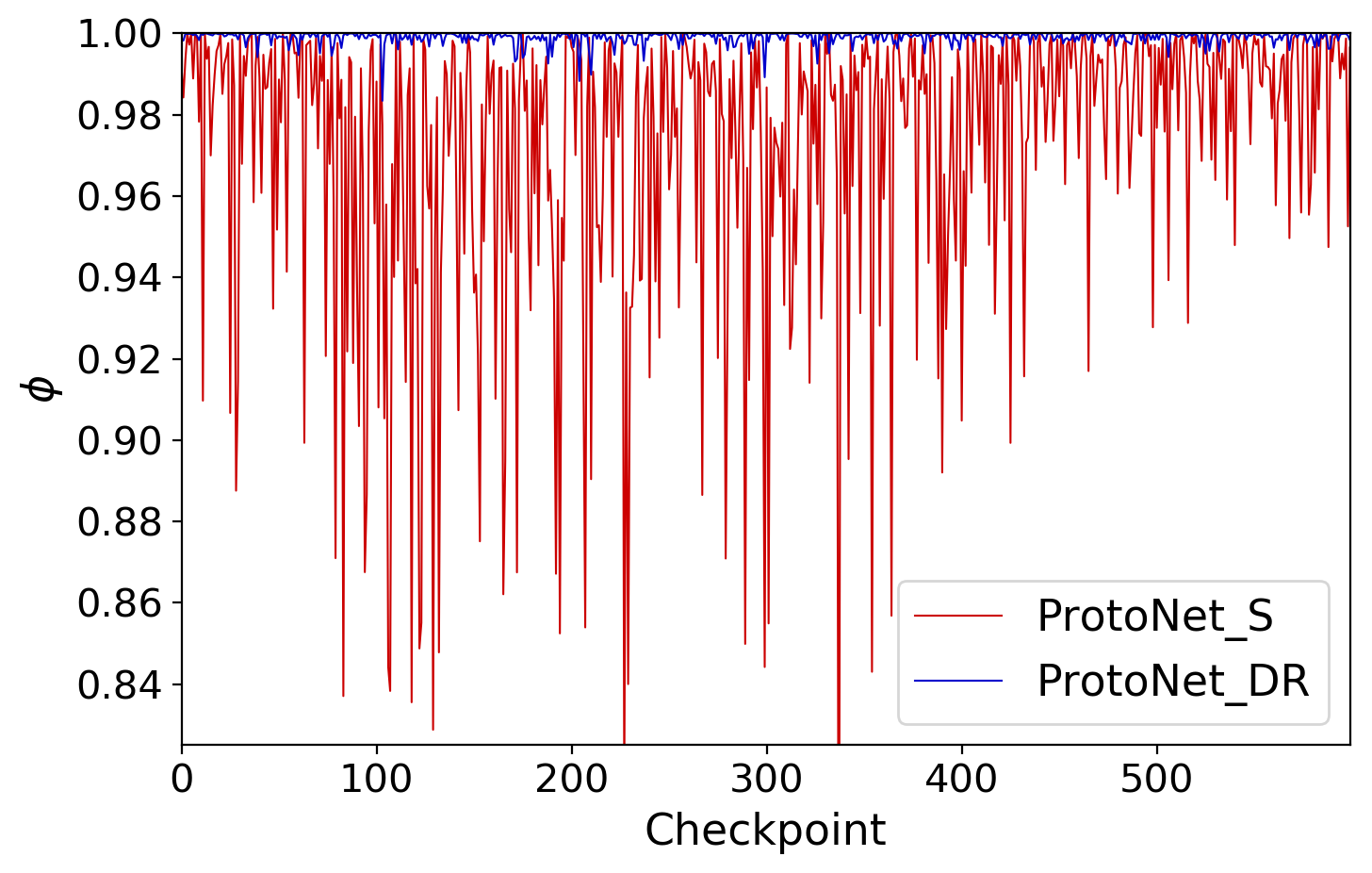} %
        \caption{Data: CUB, Backbone: Conv4}
    \end{subfigure}
    \begin{subfigure}[b]{0.24\textwidth}%
        \includegraphics[width=1.0\linewidth,height=0.625\linewidth]{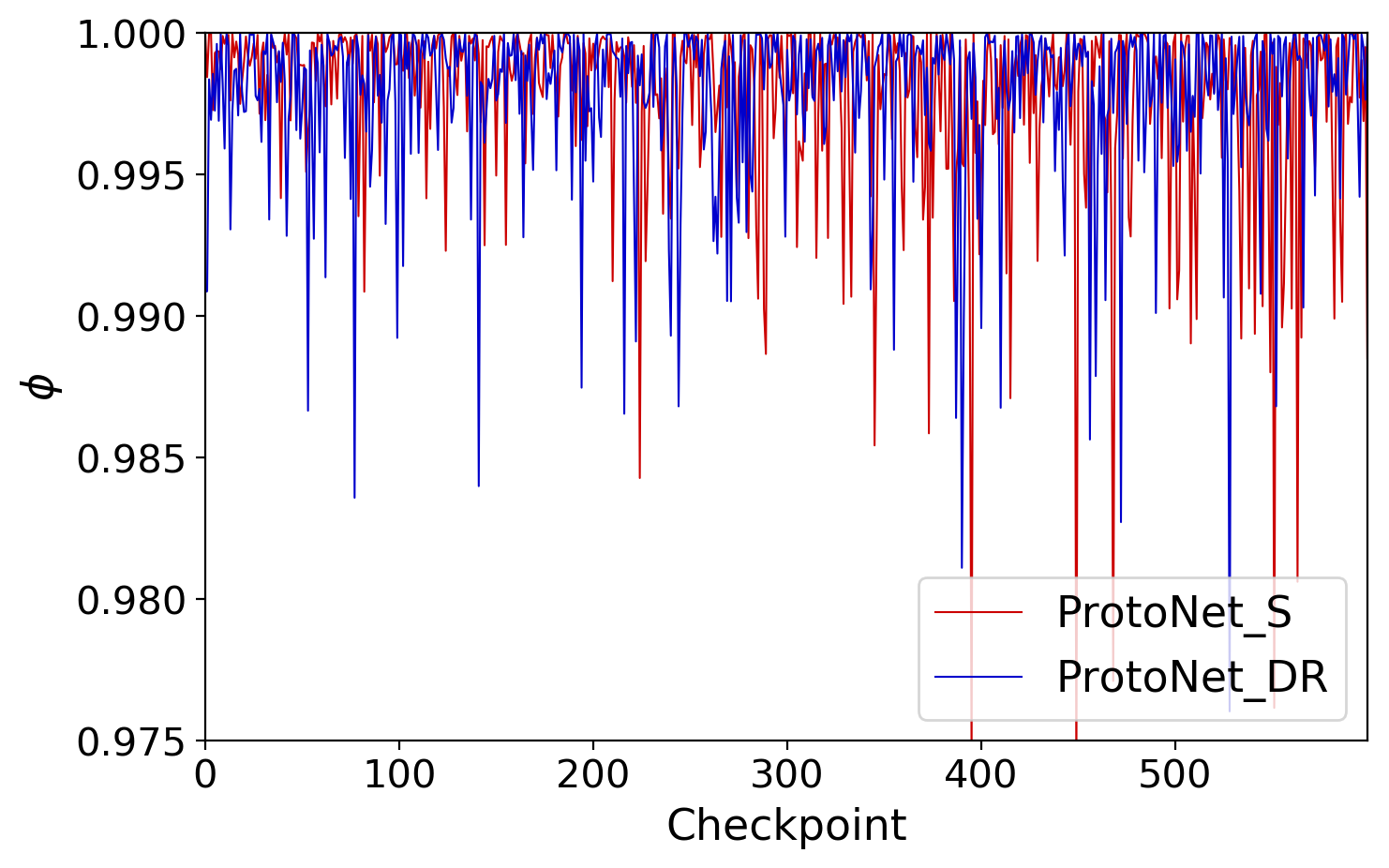} %
        \caption{Data: CUB, Backbone: ResNet18%
        }
    \end{subfigure}
    \begin{subfigure}[b]{0.24\textwidth}%
        \includegraphics[width=1.0\linewidth,height=0.625\linewidth]{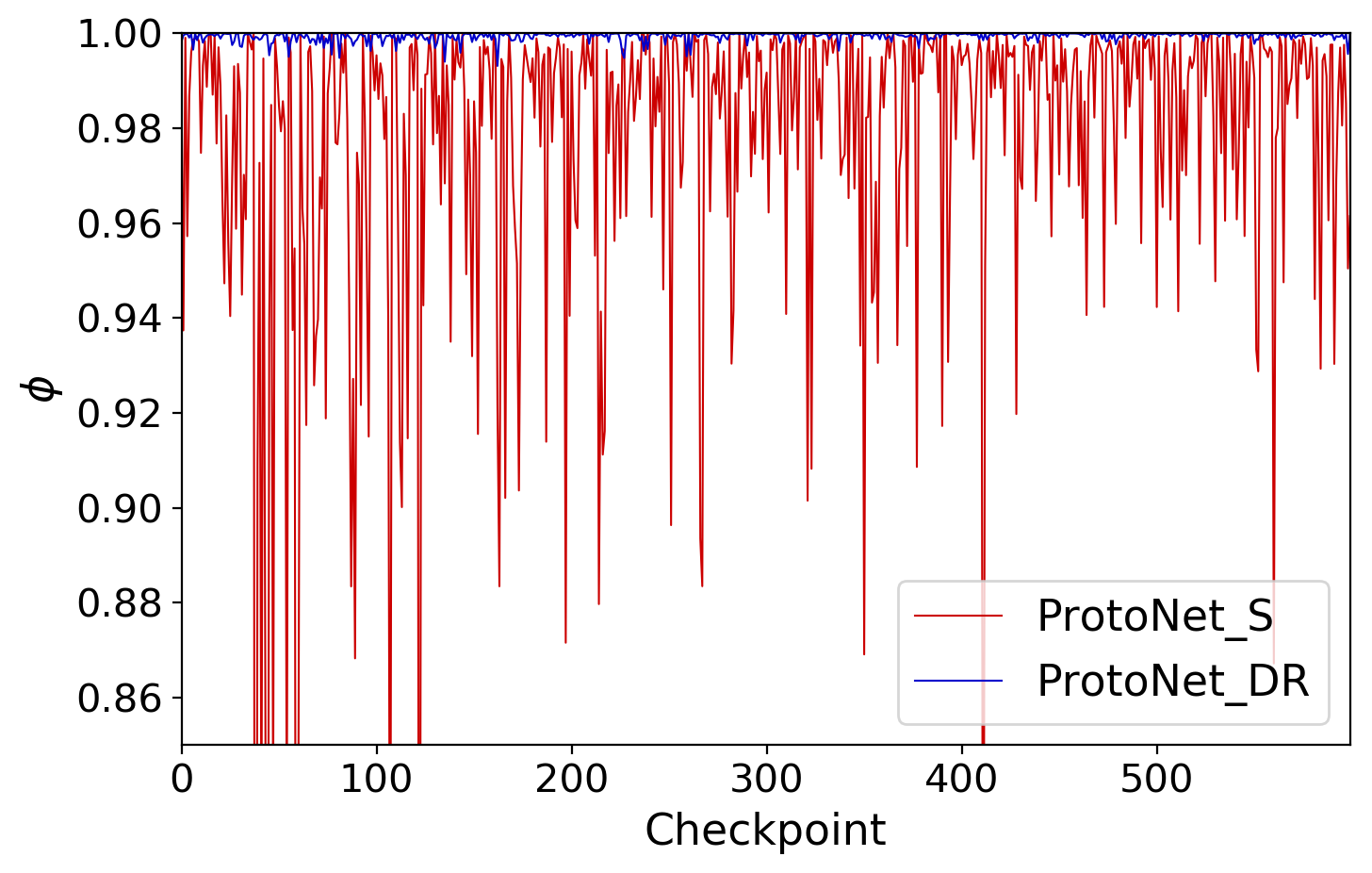} %
        \caption{Data: \emph{mini}-ImageNet, Backbone: Conv4}
    \end{subfigure}
    \begin{subfigure}[b]{0.24\textwidth}%
        \includegraphics[width=1.0\linewidth,height=0.625\linewidth]{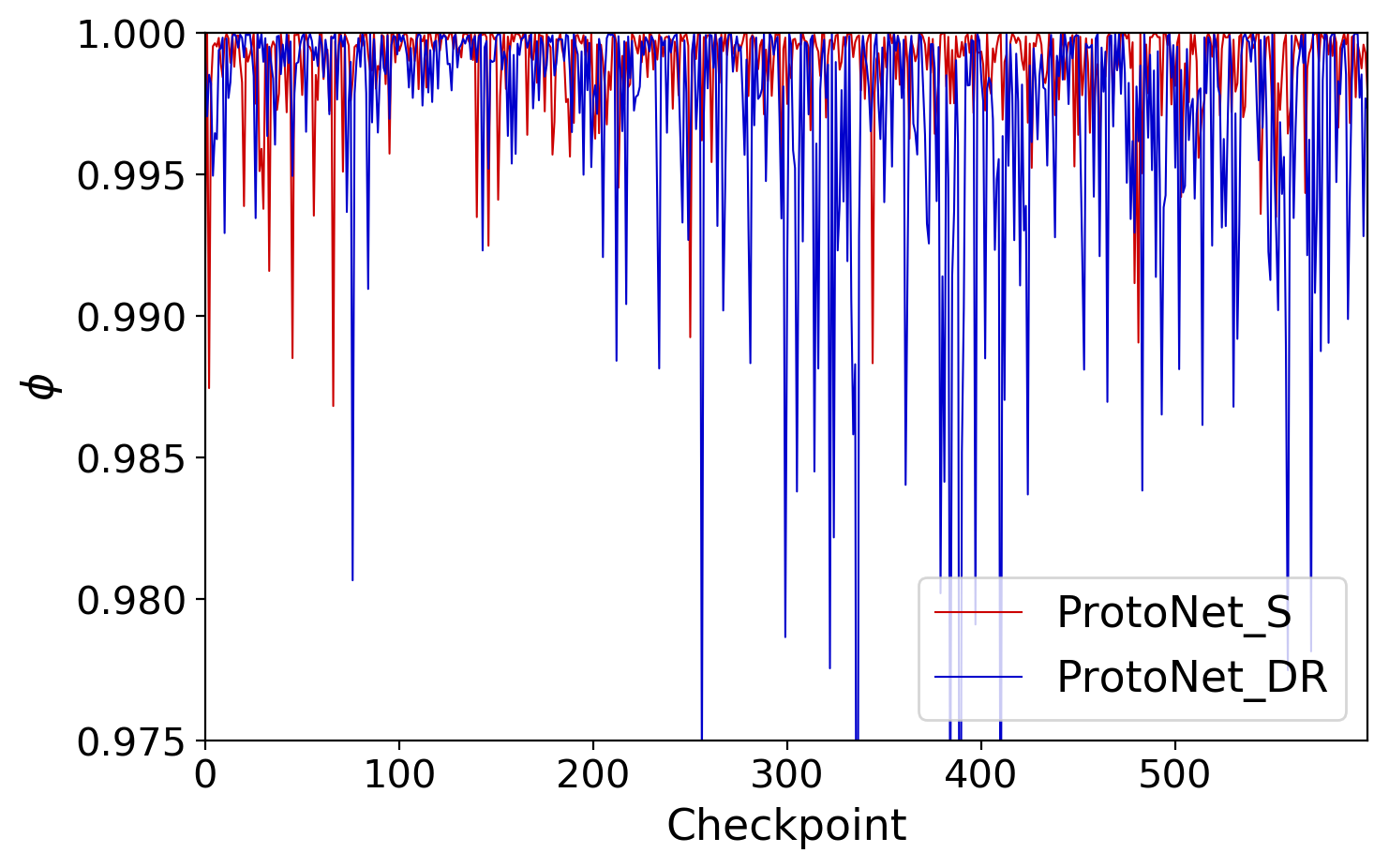} %
        \caption{Data: \emph{mini}-ImageNet, Backbone: ResNet18%
        }
    \end{subfigure}
    \caption{Norm ratio \(\phi\) curves for two different datasets and backbones on 1-shot learning tasks. Note that the ranges of the y-axis are smaller in (b) and (d) than (a) and (c).}
    \label{fig:1-shot_norm_ratio}
\end{figure*}

\begin{figure*}[t]
    \centering
    \vspace{.05in}
    \begin{subfigure}[b]{0.24\textwidth}%
        \includegraphics[width=1.0\linewidth,height=0.625\linewidth]{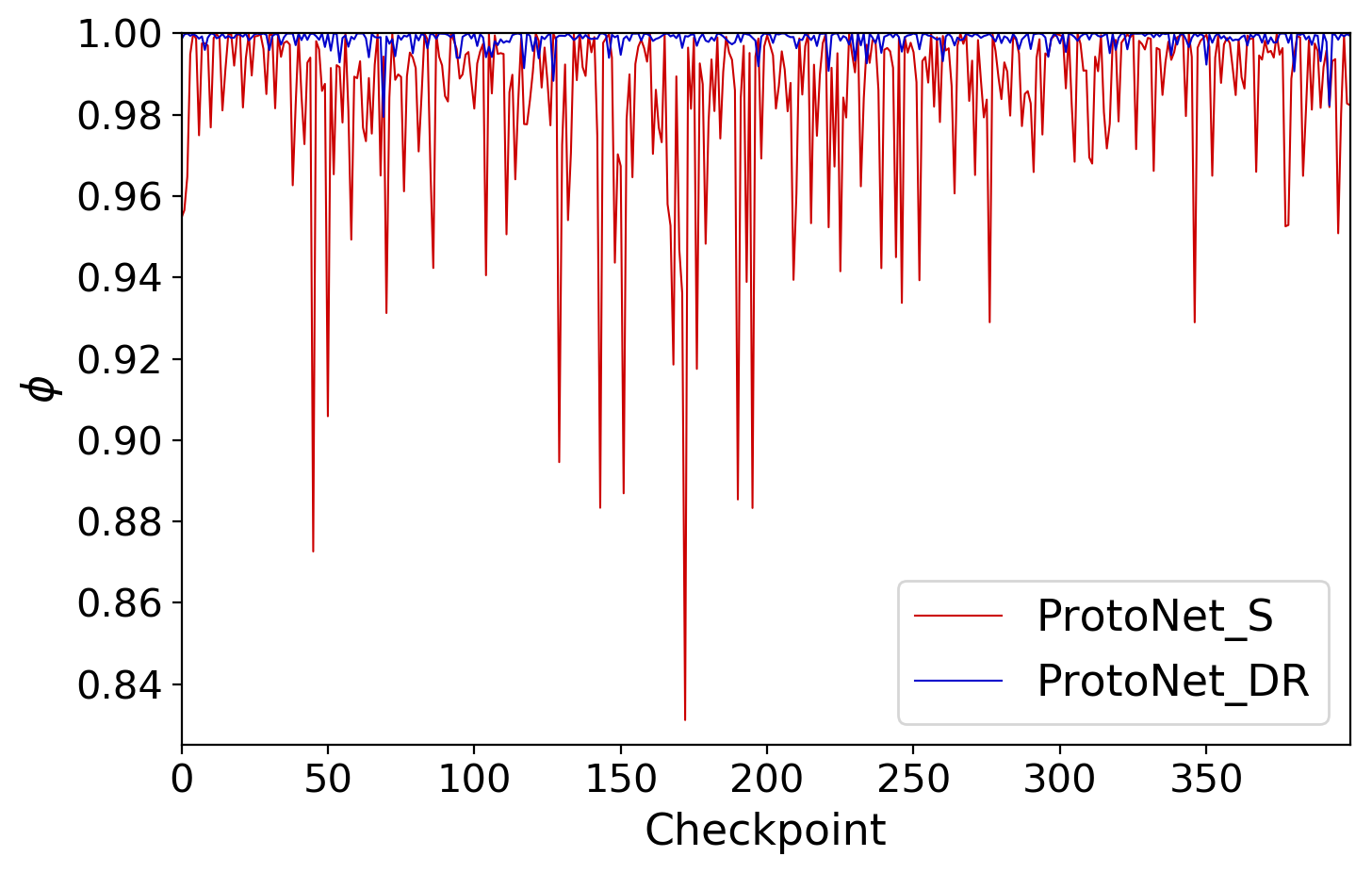} %
        \caption{Data: CUB, Backbone: Conv4}
    \end{subfigure}
    \begin{subfigure}[b]{0.24\textwidth}%
        \includegraphics[width=1.0\linewidth,height=0.625\linewidth]{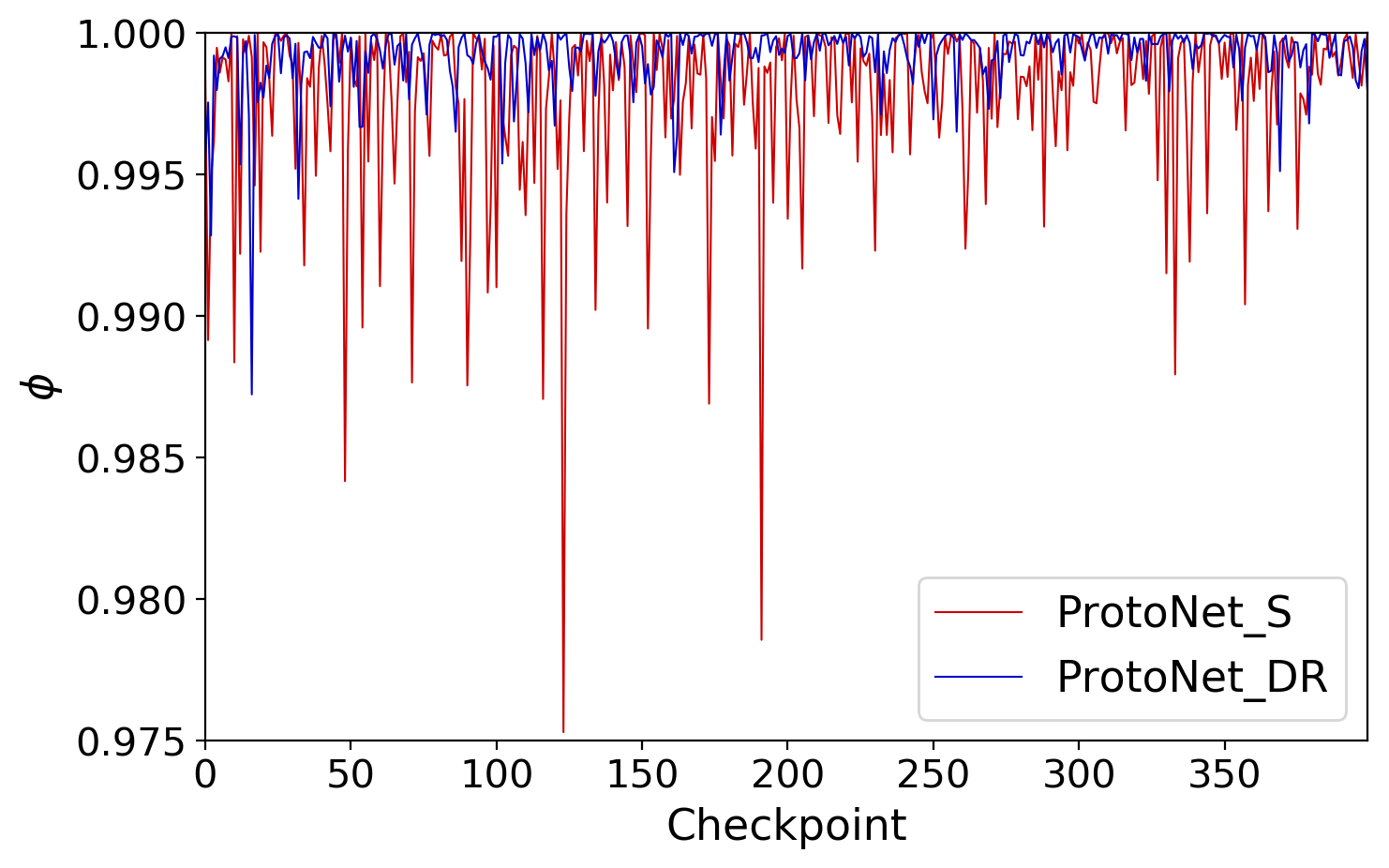} %
        \caption{Data: CUB, Backbone: ResNet18%
        }
    \end{subfigure}
    \begin{subfigure}[b]{0.24\textwidth}%
        \includegraphics[width=1.0\linewidth,height=0.625\linewidth]{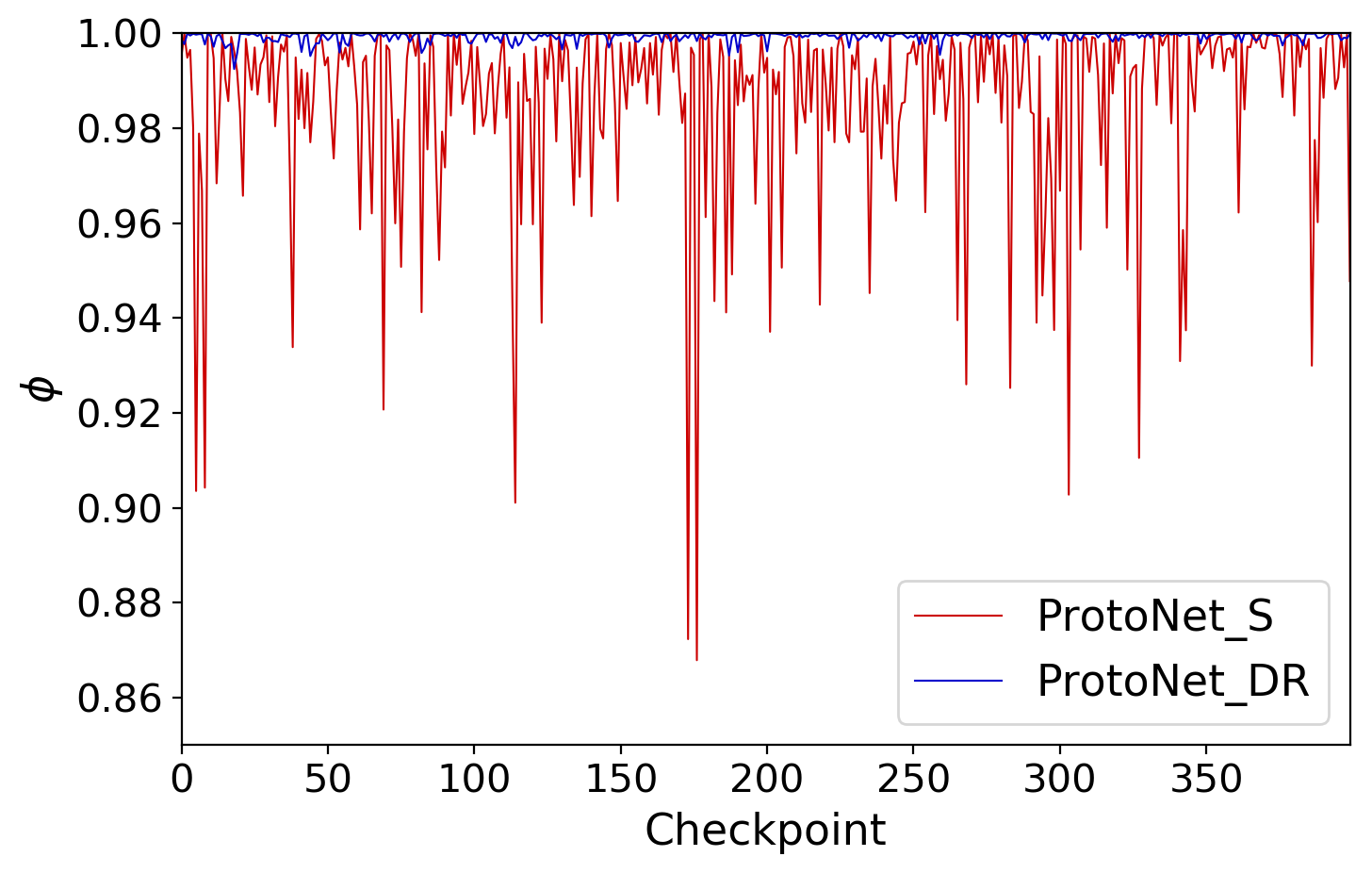} %
        \caption{Data: \emph{mini}-ImageNet, Backbone: Conv4}
    \end{subfigure}
    \begin{subfigure}[b]{0.24\textwidth}%
        \includegraphics[width=1.0\linewidth,height=0.625\linewidth]{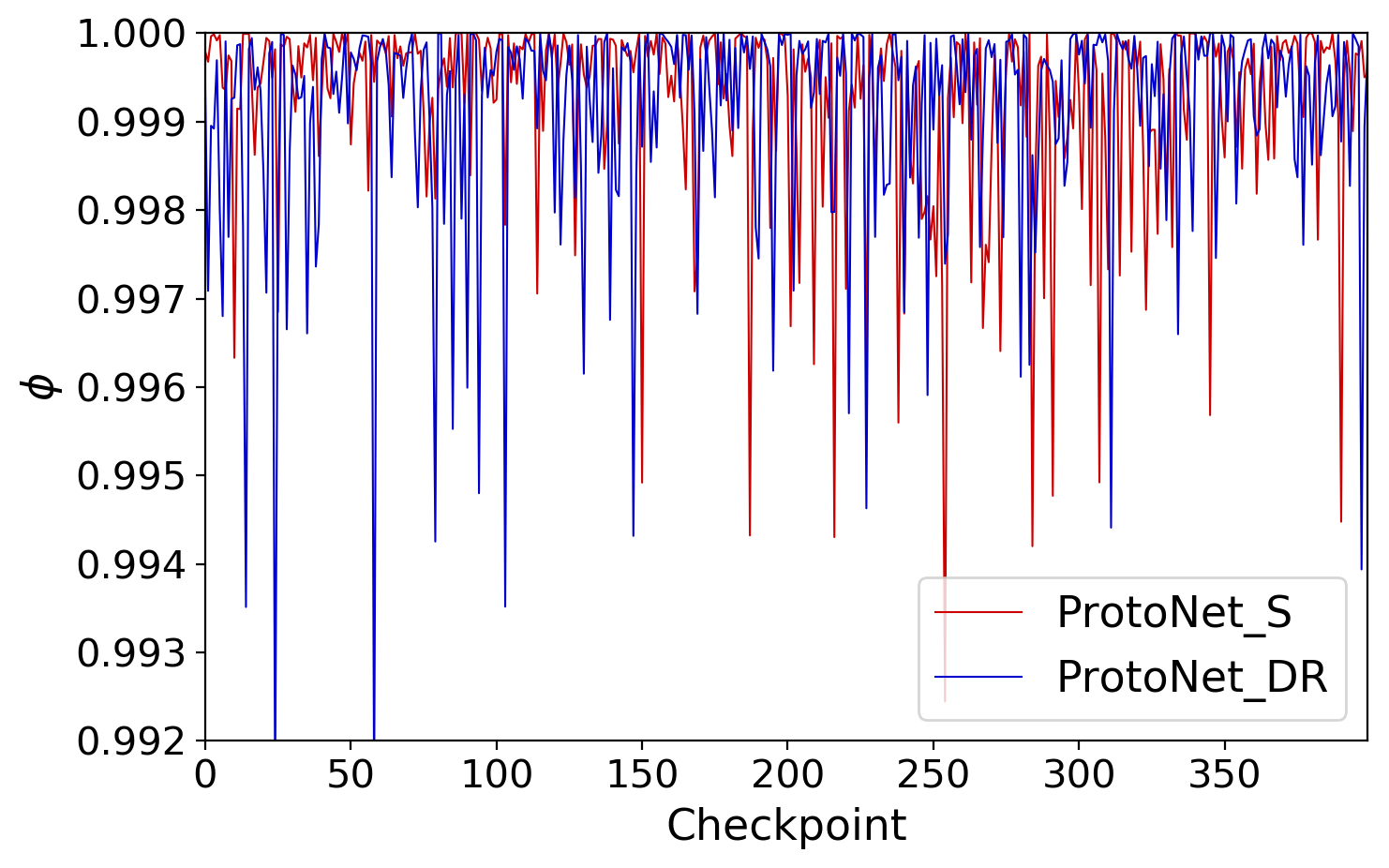} %
        \caption{Data: \emph{mini}-ImageNet, Backbone: ResNet18%
        }
    \end{subfigure}
    \caption{Norm ratio \(\phi\) curves for two different datasets and backbones on 5-shot learning tasks. Note that the ranges of the y-axis are smaller in (b) and (d) than (a) and (c).}
    \label{fig:5-shot_norm_ratio}
\end{figure*}

In addition to norm ratio \(\phi\), to analyze the location changes of query points relative to the positions of prototypes, for each checkpoint, %
we also calculated other proposed measures called \emph{con-alpha ratio}  \(\frac{\psi_{con}}{\hat{\alpha^*}}\), \emph{div-alpha ratio} \(\frac{\psi_{div}}{\hat{\alpha^*}}\), and \emph{con-div ratio} \(\frac{\psi_{con}}{\psi_{div}}\).
We use the same estimated scale parameter \(\hat{\alpha^*}\) as defined in the previous paragraph and Appendix \ref{sec:proposed_norm_ratio}. Value \(\psi_{con}\) measures a degree of convergence of query points to the prototypes with the same class. Con-alpha ratio \(\frac{\psi_{con}}{\hat{\alpha^*}}\) is the corresponding value %
after adjusting scale changes. It %
is smaller than \(1\) when query points get close to the prototypes %
with the same classes after adjusting the scale changes. %
Value \(\psi_{div}\) measures a degree of divergence (separation) of query points to the prototypes with different classes. Div-alpha ratio \(\frac{\psi_{div}}{\hat{\alpha^*}}\) is the corresponding value %
after adjusting the scale changes. It %
is larger than \(1\) when query points get far apart from the prototypes %
with different classes after adjusting the scaling. Con-div ratio \(\frac{\psi_{con}}{\psi_{div}}\) is defined as 
con-alpha ratio \(\frac{\psi_{con}}{\hat{\alpha^*}}\) divided by div-alpha ratio \(\frac{\psi_{div}}{\hat{\alpha^*}}\).
It measures a relative degree of intended convergence of query points compared to divergence to the prototypes with different classes. Detailed explanations for these measures are in Appendix \ref{sec:proposed_ratios_other_measures}. 

\subsection{Experiment Results}

Using the Conv4 backbone (Figures \ref{fig:1-shot_training_a}, \ref{fig:1-shot_training_c}, \ref{fig:5-shot_training_a}, and \ref{fig:5-shot_training_c}), we can observe that %
utilization of the DR formulation helps to train faster than using the softmax-based prototypical networks \citep{snell2017prototypical}. When using the ResNet18 backbone, the differences are smaller in 1-shot learning tasks (Figures \ref{fig:1-shot_training_b} and \ref{fig:1-shot_training_d}) or %
reversed in 5-shot tasks (Figures \ref{fig:5-shot_training_b} and \ref{fig:5-shot_training_d}). 

Figures \ref{fig:1-shot_norm_ratio} and \ref{fig:5-shot_norm_ratio} visualize the changes of norm ratio \(\phi\). Tables \ref{tab:CUB_Conv4_proposed_ratios} to \ref{tab:miniImg_ResNet18_proposed_ratios_5s} (in Appendix \ref{sec:proposed_ratios_results}) report geometric means and statistical test results on norm ratio values. %
In all the experiments with the Conv4 backbone, %
norm ratios \(\phi\) were significantly smaller 
when we used the softmax-based formulation than the DR formulation. In other words, there were more scale changes in embedding when we used the softmax-based formulation. It indicates that when we use the Conv4 backbone, using softmax-based formulation can be more prone to weaken the training process due to the unnecessary scale changes. %
When using the ResNet18 backbone, norm ratio values were very close to \(1\) (geometric mean at least \(0.99705)\) on both formulations. 

\begin{table*}%
    \caption{Few-shot classification accuracies (\%) for CUB and \emph{mini}-ImageNet datasets. Each cell reports mean accuracy based on 600 random test episodes and the corresponding 95\% confidence interval from a single trained model. }%
    \label{tab:FSL}
    \centering
    \begin{tabular}{M{2.25cm}|M{2.25cm}|\Mc{2.0cm}\Mc{2.0cm}|\Mc{2.0cm}\Mc{2.0cm}}
    \toprule
    \multirow{2}{*}{ Backbone }&\multirow{2}{*}{ Method }& \multicolumn{2}{c}{\scriptsize CUB}& \multicolumn{2}{c}{\scriptsize \emph{mini}-ImageNet} \\ \cline{3-6} 
    \scriptsize { }&\scriptsize { }&\scriptsize 1-shot &\scriptsize 5-shot & \scriptsize 1-shot &\scriptsize 5-shot \\ \midrule
    \multirow{2}{*}{ Conv4}&    %
    \scriptsize {ProtoNet\_S}&\scriptsize {\(50.46 \pm 0.88\)} &\scriptsize \(76.39 \pm 0.64\)
    &\scriptsize \(44.42 \pm 0.84\) & \scriptsize \(64.24 \pm 0.72\)\\ \cline{2-6} 
    {}&\scriptsize { ProtoNet\_{DR}}&\scriptsize {\(57.13 \pm 0.95\)} &\scriptsize \(76.50 \pm 0.66\)
    &\scriptsize \( 48.71 \pm 0.78\) & \scriptsize \( 65.90\pm 0.69\)\\ \hline
    \multirow{2}{*}{ \makecell{ResNet18} }&
    \scriptsize {ProtoNet\_S}&\scriptsize {\(72.99 \pm 0.88 \)} &\scriptsize \(86.64 \pm 0.51\)
    &\scriptsize \(54.16 \pm 0.82\) & \scriptsize \(73.68 \pm 0.65\)\\ \cline{2-6} 
    {}&\scriptsize { ProtoNet\_{DR}}&\scriptsize {\(73.33 \pm 0.90\)} &\scriptsize \(86.63 \pm 0.49\)
    &\scriptsize \( 54.86 \pm 0.86\) & \scriptsize \( 72.93\pm 0.64\)\\ 
    \bottomrule
    \end{tabular}
\end{table*}

Tables \ref{tab:CUB_Conv4_proposed_ratios} to \ref{tab:miniImg_ResNet18_proposed_ratios_5s} (in Appendix \ref{sec:proposed_ratios_results}) also report geometric means, proportions of properly learned cases (\(\frac{\psi_{con}}{\hat{\alpha^*}}<1\), \(\frac{\psi_{div}}{\hat{\alpha^*}}>1\), \(\frac{\psi_{con}}{\psi_{div}}<1\)), and statistical test results on con-alpha ratio \(\frac{\psi_{con}}{\hat{\alpha^*}}\), div-alpha ratio \(\frac{\psi_{div}}{\hat{\alpha^*}}\), and con-div ratio \(\frac{\psi_{con}}{\psi_{div}}\). %
When we consider con-alpha ratio \(\frac{\psi_{con}}{\hat{\alpha^*}}\) for 1-shot learning tasks, the %
properly converged cases (\(\frac{\psi_{con}}{\hat{\alpha^*}}<1\)) were significantly more frequent when we used DR formulation. It means that in 1-shot training, our DR formulation model is more stable in decreasing distances between query points and prototypes with the corresponding classes. When we consider div-alpha ratio \(\frac{\psi_{div}}{\hat{\alpha^*}}\), for all the experiments, the properly diverged cases (\(\frac{\psi_{div}}{\hat{\alpha^*}}>1\)) were significantly more frequent when we used the DR formulation.
It indicates that the DR formulation-based model is more stable in increasing the distance between query points and prototypes with different classes.

Table \ref{tab:FSL} reports few-shot classification accuracies on test episodes. First, we consider the results when the Conv4 backbone was used for training. Except for the 5-shot classification task on the CUB dataset, which resulted in comparable accuracies (difference was \(0.11\%\)), the test accuracies were higher with the DR formulation. The accuracy differences ranged from \(1.66\%\) to \(6.67\%\). 

Now, we consider the results with the ResNet18 backbone in Table \ref{tab:FSL}. %
First, the accuracy gaps were reduced. \citet{chen2019closer} also observed this phenomenon when they compared accuracy gaps with different backbones using several few-shot learning models.
While the differences were small in the 1-shot classification task (differences were \(0.34\%\) or \(0.70\%\)), using DR formulation achieved higher accuracies. For the 5-shot classification task on the \emph{mini}-ImageNet dataset, using DR formulation achieved \(0.75\%\) lower accuracy. 

\section{Discussion}

In this work, we address the limitations of softmax-based formulation for metric learning by proposing a distance-ratio-based (DR) formulation. DR formulation focuses on updating relative positions on embedding by ignoring the scale of an embedding space. It also enables stable training by using each representing point as an optimal position. %
Our experiments show that using DR formulation resulted in faster training speed in general and improved or comparable generalization performances.

When distance \(d_{x',c}\) is a distance between a query point \(x'\) and the nearest support point with class \(c\), distance ratio \(\frac{d_{x',c_1}}{d_{x',c_2}}\) for two different classes \(c_1\) and \(c_2\) has been utilized in previous literature. %
By setting \(c_1\) as the nearest class and \(c_2\) as the second nearest class from a query point \(x'\), \citet{junior2017nearest} have used the distance ratio named {\it nearest neighbor distance ratio (NNDR)} for handling open-set classification problems \citep{geng2020recent}.
Independently, \citet{jiang2018trust} have utilized the inverse value \(\frac{d_{x',c_2}}{d_{x',c_1}}\) %
to define {\it trust score}, which is an alternative value for confidence value of the %
standard softmax classifiers. Unlike these works that directly use distance-ratios without modeling confidence values, %
distance-ratio-based (DR) formulation models probability \(p(y=c|x')\) using distance ratios. 

To output %
confidence scores that are either \(0\) or \(1\) on some areas,
sparsemax \citep{martins2016softmax}, sparsegenlin, and sparsehourglass \citep{laha2018controllable}
were proposed as alternatives for softmax formulation in non-metric learning cases. Unlike DR formulation, which has %
a confidence score \(0\) or \(1\) only on (countable) \emph{%
representing points}, these formulations have \emph{areas} (sets of uncountable points) %
that output confidence score \(0\) or \(1\). Such property is inappropriate for metric learning %
as confidence scores can be \(1\) even for %
non-representing points, and thus query points do not need to converge very close to the corresponding representing points.

Recent works \citep{wang2017normface, liu2017sphereface, wang2018additive, wang2018cosface, deng2019arcface} proposed to use modifications of the standard softmax classifier for metric learning. These modified softmax classifiers showed competitive performances on metric learning \citep{musgrave2020metric}. %
Unlike traditional metric learning models that use data points or prototypes, which are obtained from data points, to represent classes, they use learnable parameter vectors to represent classes. They %
use cosine similarity %
on normalized embedding space \(\mathbb{S}^{\left(d_{F}-1\right)}\). %
That is equivalent to using the softmax-based formulation with Euclidean distance. %
While scale dependency of the softmax-based formulation can be addressed due to normalization, the softmax-based formulation still lacks the second property of DR formulation. 
Thus, %
a representing parameter vector may not be a vector that maximizes the confidence value. %
To handle this issue, DR formulation can also be used as an alternative %
by using Euclidean or angular distance on normalized embedding space (see example in Appendix \ref{sec:normalized_ex}).%

In addition to the supervised metric learning, cosine similarity on a normalized representation space is also used in %
contrastive self-supervised learning \citep{chen2020simple} and in recent data augmentation strategy \citep{khosla2020supervised} which uses supervised contrastive loss. DR formulation can also be applied to these models for possible performance improvements. 

While using DR formulation resulted in faster or comparable training speed in most experiments, in Figure \ref{fig:5-shot_training_b} and \ref{fig:5-shot_training_d}, we observe slightly slower training speed in 5-shot learning with the ResNet18 backbone. We speculate the reason is %
that an average point is not an optimal point to represent a class in DR formulation (explained in Appendix \ref{sec:represent_mean_DR}), unlike softmax-based formulation (explained in Appendix \ref{sec:represent_mean_softmax}). To investigate this, in Appendix \ref{sec:1-NN_5s}, %
we conduct 5-shot learning experiments with nearest neighbor-based models instead of using an average point to represent a class. %

Our experiment results showed that the scale changes of softmax-based prototypical networks are decreased dramatically when the ResNet18 was used as a backbone. %
One possible reason for this phenomenon can be %
the skip connections in residual modules \citep{he2016deep} and the fact that the scale of the input layer is fixed. It requires further investigation to draw a conclusion.

\bibliography{My_library} %
\bibliographystyle{iclr2021_conference}

\onecolumn
\renewcommand\thesection{\Alph{section}}

\appendix
\section*{APPENDIX}

\section{Using an Average Point to Represent a Class}\label{sec:represent_mean}

Here, we consider what is an appropriate point to represent a class \(c\) based on given support points. We denote a set of support points with class \(c\) as \(S_{c}\).

\subsection{Softmax-Based Formulation}\label{sec:represent_mean_softmax}

The softmax-based formulation can be considered as an estimation of probability \(p(y=c|x')\) when Gaussian distribution is used to estimate class-conditional distribution \(p(x'|y=c)\). Mathematically, we consider an estimation \(\hat{p}(x'|y=c)\) as:
\begin{flalign}
\hat{p}(x'|y=c)=\frac{1}{\sigma \sqrt{2\pi}} \exp{ \left(-\frac{1}{2} \left(\frac{d(f_{\theta}(x'),\mathbf{r}_{c}))}{\sigma}\right)^{2} \right)},\nonumber
\end{flalign}
where \(\sigma\) is a standard deviation of the distribution and \(\mathbf{r}_{c}\) is a representing point of class \(c\). When we fix \(\sigma=\frac{1}{\sqrt{2}}\), we get a simpler equation:
\begin{flalign}
\hat{p}(x'|y=c)=\frac{1}{\sqrt{\pi}} \exp{ \left(- d(f_{\theta}(x'),\mathbf{r}_{c}))^{2} \right)}\nonumber
\end{flalign}

Based on this equation and support points, %
we use MLE (maximum likelihood estimation) approach to find a good estimation for representing point \(\mathbf{r}_{c}\). %
To maximize the likelihood, let us denote the corresponding likelihood function as \(L(\mathbf{r}_{c};S_{c})\). It satisfies equations:
\begin{flalign}
L(\mathbf{r}_{c};S_{c})=&\prod_{(x_i,y_i)\in S_c}{\hat{p}(x_i|y=y_i)}\nonumber\\
=&\frac{1}{\sqrt{\pi}^{|S_c|}} \exp{\left( -\sum\limits_{(x_i,y_i)\in S_c} { d(f_{\theta}(x_i),\mathbf{r}_{c}))^{2}} \right)}\nonumber
\end{flalign}

As the sum \(\sum\limits_{(x_i,y_i)\in S_c} { d(f_{\theta}(x_i),\mathbf{r}_{c}))^{2}}\) is minimized by the average points of support points on embedding space, \(\mathbf{r}_{c}=\mathbf{p}_{c}=\frac{1}{|S_{c}|}\sum\limits_{(x_i,y_i)\in S_c}{f_{\theta}(x_i)}\) becomes MLE solution. %

\subsection{Distance-Ratio-Based (DR) Formulation}\label{sec:represent_mean_DR}

For special case \(\rho=2\), DR formulation can be considered as a limit distribution of \(p(y=c|x')\) when Cauchy distribution is used to estimate class-conditional distribution \(p(x'|y=c)\). Mathematically, we consider equation \citep{sebe2002emotion}: 
\begin{flalign}
\hat{p}(x'|y=c)=\frac{\gamma}{\pi\left(\gamma^2+d(f_{\theta}(x'),\mathbf{r}_{c})^2\right)}, \nonumber
\end{flalign}
where \(\gamma\) is a parameter of the distribution and \(\mathbf{r}_{c}\) is a representing point of class \(c\).
Then, when \(\gamma\) approaches \(0\) with uniform probability for \(\hat{p}(y=c)\), the limit distribution of \(\hat{p}(y=c|x')\) (defined using Bayes theorem) becomes the DR formulation. 

\newpage
When we denote the corresponding likelihood function as \(L(\mathbf{r}_{c};S_{c})\), it satisfies the equation:
\begin{flalign}
L(\mathbf{r}_{c};S_{c})%
=\left(\frac{\gamma}{\pi}\right)^{|S_c|}\frac{1}{\prod\limits_{(x_i,y_i)\in S_c}{\left(\gamma^2+d(f_{\theta}(x_i),\mathbf{r}_{c})^2\right)}}\nonumber%
\end{flalign}

For MLE, when \(\gamma\) is very close to \(0\), we need to minimize product defined as:
\begin{flalign}
\prod\limits_{(x_i,y_i)\in S_c}{d(f_{\theta}(x_i),\mathbf{r}_{c})^2}=\left(\prod\limits_{(x_i,y_i)\in S_c}{d(f_{\theta}(x_i),\mathbf{r}_{c})}\right)^2\nonumber
\end{flalign}

The points that minimize the product are \(\mathbf{r}_{c}=f_{\theta}(x_i)\) for \((x_i,y_i)\in S_c\). It means that when we consider \(K\)-shot classification with \(K>1\), using an average point would not be an optimal point to represent a class.

\section{Proof of Equations (7) %
and (8)%
}\label{sec:sparsity_proof}

\paragraph{Property} When \(d(\mathbf{p}_{c},\mathbf{p}_{c'})>0\)  %
for \(\forall c'\in {\mathcal{Y}_E}\) with \(c'\neq c\), then the following two equations hold:
\begin{flalign}
\lim_{x'\rightarrow\mathbf{p}_{c}}{\delta_{c}(x')}=1 \qquad \qquad\tag{7}\nonumber\\
\lim_{x'\rightarrow\mathbf{p}_{c'}}{\delta_{c}(x')}=0 \qquad \qquad\tag{8}\nonumber
\end{flalign}

\begin{proof}
First, let us assume \(d(\mathbf{p}_{c},\mathbf{p}_{c'})>0\) for \(c'\neq c\).

\(d_{x',c'}=d(f_{\theta}(x'),\mathbf{p}_{c'})%
\ge|d(f_{\theta}(x'),\mathbf{p}_{c})-d(\mathbf{p}_{c},\mathbf{p}_{c'})|=|d_{x',c}-d(\mathbf{p}_{c},\mathbf{p}_{c'})|\) %
(\(\because\) The reverse triangle inequality from the triangle inequality).

\(\lim\limits_{x'\rightarrow\mathbf{p}_{c}}{d_{x',c'}}%
\ge\lim\limits_{x'\rightarrow\mathbf{p}_{c}}|d_{x',c}%
-d(\mathbf{p}_{c},\mathbf{p}_{c'})|=|\left(\lim\limits_{x'\rightarrow\mathbf{p}_{c}}d_{x',c}\right)%
-d(\mathbf{p}_{c},\mathbf{p}_{c'})|=|0-d(\mathbf{p}_{c},\mathbf{p}_{c'})|=d(\mathbf{p}_{c},\mathbf{p}_{c'})>0\).

As \(\lim\limits_{x'\rightarrow\mathbf{p}_{c}}{d_{x',c'}}>0\), we get \(\lim\limits_{x'\rightarrow\mathbf{p}_{c}}
\frac{1}{d_{x',c'}^{\rho}}%
<\infty\). 

Thus, we also get \(\lim\limits_{x'\rightarrow\mathbf{p}_{c}}\left(\sum\limits_{y\in \mathcal{Y}_E, y\neq c}{ \frac{1}{d_{x',y}^\rho} }\right)<\infty\). Let us denote the limit value as \(C_{1}\).

\begin{flalign}
\lim\limits_{x'\rightarrow\mathbf{p}_{c}}{\delta_{c}(x')}=&\lim\limits_{x'\rightarrow\mathbf{p}_{c}}{\frac{ \frac{1}{d_{x',c}^\rho} }{\sum\limits_{y\in \mathcal{Y}_E}{ \frac{1}{d_{x',y}^\rho} }}}=\lim\limits_{x'\rightarrow\mathbf{p}_{c}}{\frac{1}{1+d_{x',c}^{\rho}\sum\limits_{y\in \mathcal{Y}_E, y\neq c}{ \frac{1}{d_{x',y}^\rho} } }} %
=\frac{1}{1+\lim\limits_{x'\rightarrow\mathbf{p}_{c}} \left(d_{x',c}^{\rho}\sum\limits_{y\in \mathcal{Y}_E, y\neq c}{ \frac{1}{d_{x',y}^\rho} }\right) } \nonumber\\
=&\frac{1}{1+\lim\limits_{x'\rightarrow\mathbf{p}_{c}} d_{x',c}^{\rho}\lim\limits_{x'\rightarrow\mathbf{p}_{c}}\left(\sum\limits_{y\in \mathcal{Y}_E, y\neq c}{ \frac{1}{d_{x',y}^\rho} }\right) }\label{eq:sparse_1_eq}\\
=&\frac{1}{1+0 \times C_{1}}=1 \nonumber
\end{flalign}

We proved Equation (7). %

With a similar process as above, we get \(\lim\limits_{x'\rightarrow\mathbf{p}_{c'}}{d_{x',c}}>0\) and \(\lim\limits_{x'\rightarrow\mathbf{p}_{c'}}{d_{x',c}^{\rho}}>0\). Let us denote the later limit value as \(C_{2}\).

\(\lim\limits_{x'\rightarrow\mathbf{p}_{c'}}{ \frac{1}{d_{x',c'}^\rho} }=\lim\limits_{x'\rightarrow\mathbf{p}_{c'}}{ \left(\frac{1}{d_{x',c'}}\right)^{\rho} }=\infty\), and thus \(\lim\limits_{x'\rightarrow\mathbf{p}_{c'}}\sum\limits_{y\in \mathcal{Y}_E, y\neq c}{ \frac{1}{d_{x',y}^\rho} }=\infty\).

With a similar process as in Equation (\ref{eq:sparse_1_eq}), we get an equation:
\begin{flalign}
\lim\limits_{x'\rightarrow\mathbf{p}_{c'}}{\delta_{c}(x')}=\frac{1}{1+\lim\limits_{x'\rightarrow\mathbf{p}_{c'}} d_{x',c}^{\rho}\lim\limits_{x'\rightarrow\mathbf{p}_{c'}}\left(\sum\limits_{y\in \mathcal{Y}_E, y\neq c}{ \frac{1}{d_{x',y}^\rho} }\right)} %
=\frac{1}{1+C_2 \times \infty}=0\nonumber
\end{flalign}

We proved Equation (8). %
\end{proof}

\section{Proposed Ratios to Analyze Training Process and Their Results}\label{sec:proposed_ratios}

\subsection{Norm Ratio \texorpdfstring{$\phi$}{}}\label{sec:proposed_norm_ratio}

To check the effect of scale changes in metric learning, we introduce norm ratio \(\phi\) to measure irrelevance to scale change. %
It is based on the positions of episode points (both support and query
points) on embedding space. Let us denote the original positions of embedding outputs as matrix \(X_{origin}\) and updated positions %
as matrix \(X_{new}\). When these matrices are not mean-
centered, we center the matrices so that an average
point is located at zero. %
Then, we consider \(X_{new}\) %
as a modification of scaled original data matrix \(\alpha^* X_{origin}\) %
for an unknown scaling parameter \(\alpha^*\). Mathematically, we consider the decomposition: %
\begin{flalign}
X_{new}=\alpha^* X_{origin} + (X_{new}-\alpha^* X_{origin})\nonumber
\end{flalign}

As the scaling parameter \(\alpha^*\) is unknown, we estimate \(\alpha^*\) using Procrustes analysis \citep{gower2004procrustes}. Procrustes analysis is used for superimposing two sets of points with optimal changes. We denote an estimated value for \(\alpha^*\) as \(\hat{\alpha^*}\) and define it as: %
\begin{flalign}\label{eq:def_alpha*}
\hat{\alpha^*}=\argmin\limits_{\alpha \in \mathbb{R}} \left \|  X_{new}-\alpha X_{origin}\right \|_F ,
\end{flalign}
where \(\left \| \cdot \right \|_F\) is the Frobenius norm. 

Let us denote the Frobenius inner product as \(\left \langle \cdot, \cdot \right \rangle_F\). Then, for \(\left \| X_{origin}\right \|_F\neq 0\), we get the equations:
\begin{flalign}
\left \|  X_{new}-\alpha X_{origin}\right \|_F^2=&\left \langle X_{new}-\alpha X_{origin},X_{new}-\alpha X_{origin} \right \rangle_F \nonumber\\
=&\left \|  X_{new}\right \|_F^2+\alpha^2 \left \|  X_{origin}\right \|_F^2-2\alpha\left \langle  X_{origin},  X_{new}\right \rangle_F\nonumber\\
= \left \| X_{origin}\right \|_F^2 &\left (\alpha -\frac{\left \langle  X_{origin},  X_{new}\right \rangle_F}{\left \| X_{origin}\right \|_F^2}  \right )^2+\left \| X_{new}\right \|_F^2-\frac{\left \langle  X_{origin},  X_{new}\right \rangle_F^2}{\left \| X_{origin}\right \|_F^2}\nonumber
\end{flalign}

Thus, we get \(\hat{\alpha^*}=\frac{\left \langle  X_{origin},  X_{new}\right \rangle_F}{\left \| X_{origin}\right \|_F^2} =\frac{\Tr{(X_{origin}^T X_{new})}}{\left \|X_{origin}\right \|_F^2}\) for \(\left \| X_{origin}\right \|_F\neq 0\).

Once we get an estimated \(\alpha^*\), norm ratio \(\phi\) is defined as: %
\begin{flalign}
\phi = \frac{\left \|  X_{new}-\hat{\alpha^*} X_{origin}\right \|_F}{\left \| X_{new}-X_{origin} \right \|_F}\nonumber
\end{flalign}

Norm ratio \(\phi\) is at most %
one because of our definition of \(\hat{\alpha^*}\) in Equation (\ref{eq:def_alpha*}). 
Thus, \(0 \le\phi\le 1\). 

When major changes are due to scaling of an embedding space, \(X_{new} \approx \hat{\alpha^*} X_{origin}\), %
and thus \(\phi\approx 0\). On the other hand, when changes are irrelevant to scaling, \(X_{new}-\hat{\alpha^*} X_{origin} \approx X_{new}-X_{origin}\), and thus \(\phi\approx 1\).

\subsection{Con-Alpha Ratio  \texorpdfstring{$\frac{\psi_{con}}{\hat{\alpha^*}}$}{p con/a}, Div-Alpha Ratio \texorpdfstring{$\frac{\psi_{div}}{\hat{\alpha^*}}$}{p div/a}, and Con-Div Ratio \texorpdfstring{$\frac{\psi_{con}}{\psi_{div}}$}{p con/pdiv}}\label{sec:proposed_ratios_other_measures}

Inspired by the rate of convergence in numerical analysis, we also introduce more measures to analyze the training processes. As we are using episode training with the prototypical network \citep{snell2017prototypical}, we expect a query point \(x'\) with class \(c\) to get closer to the prototype \(\mathbf{p}_{c}\) on the embedding space. %
Similarly, we expect %
a query point \(x'\) to get far apart from prototypes %
with different classes (\(\mathbf{p}_{c'}\)). To measure the degree (speed) of convergence or divergence with prototypes, %
we denote the original parameters of the embedding function as \(\theta_{origin}\) and updated parameters as \(\theta_{new}\). Then, \(f_{\theta_{origin}}\) represents the original embedding function, and \(f_{\theta_{new}}\) represents the updated embedding function. 
We denote a query point with an index \(j\) as \(x_{j}'\).
We use the following value \(\psi_{y,j}\) to check if a query point gets closer to or %
far away from a prototype \(\mathbf{p}_{y}\). It is defined as: %
\begin{flalign}
\psi_{y,j}=\frac{d(f_{\theta_{new}}(x_{j}'),\mathbf{p}_{y})}{d(f_{\theta_{origin}}(x_{j}'),\mathbf{p}_{y})}\nonumber
\end{flalign}

Case \(\psi_{y,j}<1\) means that the \(j\) th query point gets closer to the prototype \(\mathbf{p}_{y}\) for \(y\in\mathcal{Y}_{E}\). %
Case \(\psi_{y,j}>1\) means that the \(j\) th query point gets far apart from the prototype \(\mathbf{p}_{y}\). %

Based on \(\psi_{y,j}\) values, we can estimate an average degree of convergence to the corresponding class (denoted as \(\psi_{con}\)) and an average speed of divergence from different classes (denoted as \(\psi_{div}\)). These values are calculated by taking geometric means of \(\psi_{y,j}\) values. Mathematically, we define \(\psi_{con}\) and \(\psi_{div}\) as: 
\begin{flalign}
\psi_{con}=\left( \prod_{y=y_j',y\in\mathcal{Y}_{E}}{\psi_{y,j}} \right)^{\frac{1}{|I_{\text{same}}|}},\nonumber\\
\psi_{div}=\left( \prod_{y\neq y_j', y\in\mathcal{Y}_{E}}{\psi_{y,j}} \right)^{\frac{1}{|I_{\text{diff}}|}},\nonumber
\end{flalign}
where %
\(y_j'\) is the class of \(j\) th query point, \(I_{\text{same}}\) is the set of %
prototype-query index pairs with same classes, and \(I_{\text{diff}}\) is the set of %
prototype-query index pairs with different classes. %

As these values can also be affected by scaling, we use normalized values using the estimated \(\alpha^{*}\). We call the corresponding values \(\frac{\psi_{con}}{\hat{\alpha^*}}\) and \(\frac{\psi_{div}}{\hat{\alpha^*}}\) as con-alpha ratio and div-alpha ratio, respectively. Con-alpha ratio \(\frac{\psi_{con}}{\hat{\alpha^*}}\) represents a degree of convergence of query points to the prototypes with the same class after adjusting scale changes. When query points get close to %
prototypes with the same classes (compared to the scale change), on average, con-alpha ratio \(\frac{\psi_{con}}{\hat{\alpha^*}}\) will be smaller than \(1\). Div-alpha ratio \(\frac{\psi_{div}}{\hat{\alpha^*}}\) represents a degree of divergence of query points to the prototypes with different classes after adjusting scale changes. When query points get far apart from %
prototypes with the other classes (compared to the scale change), on average, the div-alpha ratio \(\frac{\psi_{div}}{\hat{\alpha^*}}\) will be larger than \(1\). When we 
divide con-alpha ratio \(\frac{\psi_{con}}{\hat{\alpha^*}}\) with div-alpha ratio \(\frac{\psi_{div}}{\hat{\alpha^*}}\), we get another value \(\frac{\psi_{con}}{\psi_{div}}\) called con-div ratio. It measures a relative degree of convergence of query points to prototypes with the same class compared to divergence to the prototypes with different classes. We expect the con-div ratio \(\frac{\psi_{con}}{\psi_{div}}\) to be smaller than \(1\) for appropriate training.

\subsection{Results of Proposed Measures}\label{sec:proposed_ratios_results}

Table \ref{tab:CUB_Conv4_proposed_ratios} to %
\ref{tab:miniImg_ResNet18_proposed_ratios_5s} show the analysis results of norm ratio \(\psi\), con-alpha ratio \(\frac{\psi_{con}}{\hat{\alpha^*}}\), div-alpha ratio \(\frac{\psi_{div}}{\hat{\alpha^*}}\), and con-div ratio \(\frac{\psi_{con}}{\psi_{div}}\) on few-shot classification tasks. Results are based on only one experiment for each formulation. For each experiment, statistical tests are applied using %
\(600\) training checkpoints for 1-shot learning or \(400\) training checkpoints for 5-shot learning. %
Mann–Whitney U test \citep{mann1947test} is applied %
to check if the obtained values using two formulations are significantly different. %
We report the proportions of properly learned counts %
(properly learned cases: con-alpha ratio \(\frac{\psi_{con}}{\hat{\alpha^*}}<1\), div-alpha ratio \(\frac{\psi_{div}}{\hat{\alpha^*}}>1\), and con-div ratio \(\frac{\psi_{con}}{\psi_{div}}<1\)). Fisher's exact test is applied to check if the obtained counts using two formulations are significantly different. %
In tables, significant p-values (\(<0.01\)) are written in bold text.

\begin{table*}[bp]%
    \caption{Analysis results of norm ratio, con-alpha ratio, div-alpha ratio, and con-div ratio on 1-shot task for CUB data with Conv4 backbone. }%
    \label{tab:CUB_Conv4_proposed_ratios}
    \centering
    \begin{tabular}{M{2.65cm}|\Mc{1.25cm}\Mc{1.25cm}|\Mc{1.25cm}\Mc{1.25cm}|\Mc{1.75cm}|\Mc{1.75cm}}
    \toprule
    \scriptsize Measures & \multicolumn{2}{c}{\scriptsize ProtoNet\_S}& \multicolumn{2}{c}{\scriptsize ProtoNet\_{DR}}& \scriptsize Mann–Whitney U test& \scriptsize Fisher's exact test \\ 
    \scriptsize { }&\scriptsize Geometric mean &\scriptsize %
    Proportion & \scriptsize Geometric mean &\scriptsize %
    Proportion & & \\ \midrule
    \scriptsize {Norm ratio \(\phi\)}&\scriptsize {\(0.97218\)} &\scriptsize {\( \)} & \scriptsize \(0.99898\) & \scriptsize \( \) & \scriptsize \( \mathbf{<10^{-120}}\) & \scriptsize \( \) \\  \hline
    \scriptsize {Con-alpha ratio \(\frac{\psi_{con}}{\hat{\alpha^*}}\)}&\scriptsize {\(0.99710\)} &\scriptsize \( 0.87000\)
    &\scriptsize \(0.99762\) & \scriptsize \( 0.95667\)& \scriptsize \( \mathbf{0.00244}\) & \scriptsize \( \mathbf{<10^{-7}}\) \\ \hline
    \scriptsize {Div-alpha ratio \(\frac{\psi_{div}}{\hat{\alpha^*}}\)}&\scriptsize {\(1.00060 \)} &\scriptsize \( 0.63667\)
    &\scriptsize \(1.00076\) & \scriptsize \( 0.80833\)& \scriptsize \( 0.61722\) & \scriptsize \( \mathbf{<10^{-10}}\) \\ \hline
    \scriptsize {Con-div ratio \(\frac{\psi_{con}}{\psi_{div}}\)}&\scriptsize {\(0.99650\)} &\scriptsize \( 0.96833\)
    &\scriptsize \(0.99687\) & \scriptsize \( 0.99500\)& \scriptsize \( 0.05251\) & \scriptsize \( \mathbf{0.00077}\) \\ %
    \bottomrule
    \end{tabular}
\end{table*}

\begin{table*}%
    \caption{Analysis results of norm ratio, con-alpha ratio, div-alpha ratio, and con-div ratio on 1-shot task for CUB data with ResNet18 backbone. %
    }%
    \label{tab:CUB_resnet18_proposed_ratios}
    \centering
    \begin{tabular}{M{2.65cm}|\Mc{1.25cm}\Mc{1.25cm}|\Mc{1.25cm}\Mc{1.25cm}|\Mc{1.75cm}|\Mc{1.75cm}}
    \toprule
    \scriptsize Measures & \multicolumn{2}{c}{\scriptsize ProtoNet\_S}& \multicolumn{2}{c}{\scriptsize ProtoNet\_{DR}}& \scriptsize Mann–Whitney U test& \scriptsize Fisher's exact test \\ 
    \scriptsize { }&\scriptsize Geometric mean &\scriptsize %
    Proportion & \scriptsize Geometric mean &\scriptsize %
    Proportion & & \\ \midrule
    \scriptsize {Norm ratio \(\phi\)}&\scriptsize {\(0.99778\)} &\scriptsize {\( \)} & \scriptsize \(0.99789\) & \scriptsize \( \) & \scriptsize \( 0.88640\) & \scriptsize \( \) \\  \hline
    \scriptsize {Con-alpha ratio \(\frac{\psi_{con}}{\hat{\alpha^*}}\)}&\scriptsize {\(0.98729\)} &\scriptsize \( 0.94833\)
    &\scriptsize \(0.98835\) & \scriptsize \( 0.99333 \)& \scriptsize \( 0.34990\) & \scriptsize \( \mathbf{<10^{-5}}\) \\ \hline
    \scriptsize {Div-alpha ratio \(\frac{\psi_{div}}{\hat{\alpha^*}}\)}&\scriptsize {\(1.00257\)} &\scriptsize \( 0.72500 \)
    &\scriptsize \(1.00247\) & \scriptsize \( 0.86000 \)& \scriptsize \( 0.11755 \) & \scriptsize \( \mathbf{<10^{-8}}\) \\ \hline
    \scriptsize {Con-div ratio \(\frac{\psi_{con}}{\psi_{div}}\)}&\scriptsize {\(0.98476\)} &\scriptsize \( 0.99500\)
    &\scriptsize \(0.98592\) & \scriptsize \( 0.99667\)& \scriptsize \(0.02294\) & \scriptsize \( 1.00000\) \\ %
    \bottomrule
    \end{tabular}
\end{table*}

\begin{table*}%
    \caption{Analysis results of norm ratio, con-alpha ratio, div-alpha ratio, and con-div ratio on 1-shot task for \emph{mini}-ImageNet data with Conv4 backbone. %
    }%
    \label{tab:miniImg_Conv4_proposed_ratios}
    \centering
    \begin{tabular}{M{2.65cm}|\Mc{1.25cm}\Mc{1.25cm}|\Mc{1.25cm}\Mc{1.25cm}|\Mc{1.75cm}|\Mc{1.75cm}}
    \toprule
    \scriptsize Measures & \multicolumn{2}{c}{\scriptsize ProtoNet\_S}& \multicolumn{2}{c}{\scriptsize ProtoNet\_{DR}}& \scriptsize Mann–Whitney U test& \scriptsize Fisher's exact test \\ 
    \scriptsize { }&\scriptsize Geometric mean &\scriptsize %
    Proportion & \scriptsize Geometric mean &\scriptsize %
    Proportion & & \\ \midrule
    \scriptsize {Norm ratio \(\phi\)}&\scriptsize {\(0.97620\)} &\scriptsize {\( \)} & \scriptsize \(0.99940\) & \scriptsize \( \) & \scriptsize \( \mathbf{<10^{-126}}\) & \scriptsize \( \) \\  \hline
    \scriptsize {Con-alpha ratio \(\frac{\psi_{con}}{\hat{\alpha^*}}\)}&\scriptsize {\(0.99776\)} &\scriptsize \( 0.85500\)
    &\scriptsize \(0.99805\) & \scriptsize \( 0.95000\)& \scriptsize \( 0.07212\) & \scriptsize \( \mathbf{<10^{-7}}\) \\ \hline
    \scriptsize {Div-alpha ratio \(\frac{\psi_{div}}{\hat{\alpha^*}}\)}&\scriptsize {\(1.00075\)} &\scriptsize \( 0.67500\)
    &\scriptsize \(1.00087\) & \scriptsize \( 0.83333\)& \scriptsize \(0.07522\) & \scriptsize \( \mathbf{<10^{-9}}\) \\ \hline
    \scriptsize {Con-div ratio \(\frac{\psi_{con}}{\psi_{div}}\)}&\scriptsize {\(0.99701\)} &\scriptsize \( 0.95333\)
    &\scriptsize \(0.99718\) & \scriptsize \( 0.99000\)& \scriptsize \(0.41342\) & \scriptsize \( \mathbf{0.00016}\) \\ %
    \bottomrule
    \end{tabular}
\end{table*}

\begin{table*}%
    \caption{Analysis results of norm ratio, con-alpha ratio, div-alpha ratio, and con-div ratio on 1-shot task for \emph{mini}-ImageNet data with ResNet18 backbone. %
    }%
    \label{tab:miniImg_ResNet18_proposed_ratios}
    \centering
    \begin{tabular}{M{2.65cm}|\Mc{1.25cm}\Mc{1.25cm}|\Mc{1.25cm}\Mc{1.25cm}|\Mc{1.75cm}|\Mc{1.75cm}}
    \toprule
    \scriptsize Measures & \multicolumn{2}{c}{\scriptsize ProtoNet\_S}& \multicolumn{2}{c}{\scriptsize ProtoNet\_{DR}}& \scriptsize Mann–Whitney U test& \scriptsize Fisher's exact test \\ 
    \scriptsize { }&\scriptsize Geometric mean &\scriptsize %
    Proportion & \scriptsize Geometric mean &\scriptsize %
    Proportion & & \\ \midrule
    \scriptsize {Norm ratio \(\phi\)}&\scriptsize {\(0.99889\)} &\scriptsize {\( \)} & \scriptsize \(0.99705\) & \scriptsize \( \) & \scriptsize \( \mathbf{<10^{-15}}\) & \scriptsize \( \) \\  \hline
    \scriptsize {Con-alpha ratio \(\frac{\psi_{con}}{\hat{\alpha^*}}\)}&\scriptsize {\(0.99194\)} &\scriptsize \( 0.95000\)
    &\scriptsize \(0.99207\) & \scriptsize \( 0.98333\)& \scriptsize \( 0.11545\) & \scriptsize \( \mathbf{0.00187}\) \\ \hline
    \scriptsize {Div-alpha ratio \(\frac{\psi_{div}}{\hat{\alpha^*}}\)}&\scriptsize {\(1.00177\)} &\scriptsize \( 0.67167\)
    &\scriptsize \(1.00179\) & \scriptsize \(0.88167 \)& \scriptsize \(0.35608\) & \scriptsize \( \mathbf{<10^{-17}}\) \\ \hline
    \scriptsize {Con-div ratio \(\frac{\psi_{con}}{\psi_{div}}\)}&\scriptsize {\(0.99019\)} &\scriptsize \( 0.99500\)
    &\scriptsize \(0.99030\) & \scriptsize \( 0.99500\)& \scriptsize \(0.68248\) & \scriptsize \( 1.00000\) \\ %
    \bottomrule
    \end{tabular}
\end{table*}

\begin{table*}%
    \caption{Analysis results of norm ratio, con-alpha ratio, div-alpha ratio, and con-div ratio on 5-shot task for CUB data with Conv4 backbone. }%
    \label{tab:CUB_Conv4_proposed_ratios_5s}
    \centering
    \begin{tabular}{M{2.65cm}|\Mc{1.25cm}\Mc{1.25cm}|\Mc{1.25cm}\Mc{1.25cm}|\Mc{1.75cm}|\Mc{1.75cm}}
    \toprule
    \scriptsize Measures & \multicolumn{2}{c}{\scriptsize ProtoNet\_S}& \multicolumn{2}{c}{\scriptsize ProtoNet\_{DR}}& \scriptsize Mann–Whitney U test& \scriptsize Fisher's exact test \\ 
    \scriptsize { }&\scriptsize Geometric mean &\scriptsize %
    Proportion & \scriptsize Geometric mean &\scriptsize %
    Proportion & & \\ \midrule
    \scriptsize {Norm ratio \(\phi\)}&\scriptsize {\(0.98439\)} &\scriptsize {\( \)} & \scriptsize \(0.99863\) & \scriptsize \( \) & \scriptsize \( \mathbf{<10^{-61}}\) & \scriptsize \( \) \\  \hline
    \scriptsize {Con-alpha ratio \(\frac{\psi_{con}}{\hat{\alpha^*}}\)}&\scriptsize {\(0.99885\)} &\scriptsize \(  0.81500\)
    &\scriptsize \(0.99906\) & \scriptsize \(  0.87250\)& \scriptsize \( <\mathbf{10^{-4}}\) & \scriptsize \( 0.03188\) \\ \hline
    \scriptsize {Div-alpha ratio \(\frac{\psi_{div}}{\hat{\alpha^*}}\)}&\scriptsize {\(1.00095 \)} &\scriptsize \( 0.70500 \)
    &\scriptsize \(1.00114\) & \scriptsize \( 0.91500 \)& \scriptsize \(0.02577\) & \scriptsize \( \mathbf{<10^{-13}}\) \\ \hline
    \scriptsize {Con-div ratio \(\frac{\psi_{con}}{\psi_{div}}\)}&\scriptsize {\(0.99791\)} &\scriptsize \( 0.92000\)
    &\scriptsize \(0.99793\) & \scriptsize \( 0.97000 \)& \scriptsize \( 0.29667\) & \scriptsize \( \mathbf{0.00281}\) \\ %
    \bottomrule
    \end{tabular}
\end{table*}

\begin{table*}%
    \caption{Analysis results of norm ratio, con-alpha ratio, div-alpha ratio, and con-div ratio on 5-shot task for CUB data with ResNet18 backbone. %
    }%
    \label{tab:CUB_resnet18_proposed_ratios_5s}
    \centering
    \begin{tabular}{M{2.65cm}|\Mc{1.25cm}\Mc{1.25cm}|\Mc{1.25cm}\Mc{1.25cm}|\Mc{1.75cm}|\Mc{1.75cm}}
    \toprule
    \scriptsize Measures & \multicolumn{2}{c}{\scriptsize ProtoNet\_S}& \multicolumn{2}{c}{\scriptsize ProtoNet\_{DR}}& \scriptsize Mann–Whitney U test& \scriptsize Fisher's exact test \\ 
    \scriptsize { }&\scriptsize Geometric mean &\scriptsize %
    Proportion & \scriptsize Geometric mean &\scriptsize %
    Proportion & & \\ \midrule
    \scriptsize {Norm ratio \(\phi\)}&\scriptsize {\(0.99790\)} &\scriptsize {\( \)} & \scriptsize \(0.99928\) & \scriptsize \( \) & \scriptsize \( \mathbf{<10^{-14}}\) & \scriptsize \( \) \\  \hline
    \scriptsize {Con-alpha ratio \(\frac{\psi_{con}}{\hat{\alpha^*}}\)}&\scriptsize {\(0.99435\)} &\scriptsize \( 0.96999\)
    &\scriptsize \(0.99482\) & \scriptsize \( 0.98250  \)& \scriptsize \(0.22264\) & \scriptsize \( 0.088458\) \\ \hline
    \scriptsize {Div-alpha ratio \(\frac{\psi_{div}}{\hat{\alpha^*}}\)}&\scriptsize {\(1.00210\)} &\scriptsize \( 0.78000\)
    &\scriptsize \(1.00202\) & \scriptsize \( 0.96750 \)& \scriptsize \( 0.05866\) & \scriptsize \( \mathbf{<10^{-15}}\) \\ \hline
    \scriptsize {Con-div ratio \(\frac{\psi_{con}}{\psi_{div}}\)}&\scriptsize {\(0.99227\)} &\scriptsize \( 0.99250\)
    &\scriptsize \(0.99281\) & \scriptsize \( 0.99500\)& \scriptsize \(0.11123\) & \scriptsize \( 1.00000\) \\ %
    \bottomrule
    \end{tabular}
\end{table*}

\begin{table*}%
    \caption{Analysis results of norm ratio, con-alpha ratio, div-alpha ratio, and con-div ratio on 5-shot task for \emph{mini}-ImageNet data with Conv4 backbone. %
    }%
    \label{tab:miniImg_Conv4_proposed_ratios_5s}
    \centering
    \begin{tabular}{M{2.65cm}|\Mc{1.25cm}\Mc{1.25cm}|\Mc{1.25cm}\Mc{1.25cm}|\Mc{1.75cm}|\Mc{1.75cm}}
    \toprule
    \scriptsize Measures & \multicolumn{2}{c}{\scriptsize ProtoNet\_S}& \multicolumn{2}{c}{\scriptsize ProtoNet\_{DR}}& \scriptsize Mann–Whitney U test& \scriptsize Fisher's exact test \\ 
    \scriptsize { }&\scriptsize Geometric mean &\scriptsize %
    Proportion & \scriptsize Geometric mean &\scriptsize %
    Proportion & & \\ \midrule
    \scriptsize {Norm ratio \(\phi\)}&\scriptsize {\(0.98565\)} &\scriptsize {\( \)} & \scriptsize \(0.99930\) & \scriptsize \( \) & \scriptsize \( \mathbf{<10^{-69}}\) & \scriptsize \( \) \\  \hline
    \scriptsize {Con-alpha ratio \(\frac{\psi_{con}}{\hat{\alpha^*}}\)}&\scriptsize {\(0.99917\)} &\scriptsize \( 0.79500\)
    &\scriptsize \(0.99934\) & \scriptsize \( 0.79500 \)& \scriptsize \( 0.04840\) & \scriptsize \( 1.0\) \\ \hline
    \scriptsize {Div-alpha ratio \(\frac{\psi_{div}}{\hat{\alpha^*}}\)}&\scriptsize {\(1.00110\)} &\scriptsize \( 0.78000 \)
    &\scriptsize \(1.00144\) & \scriptsize \( 0.96000 \)& \scriptsize \(\mathbf{0.00054}\) & \scriptsize \( \mathbf{<10^{-14}}\) \\ \hline
    \scriptsize {Con-div ratio \(\frac{\psi_{con}}{\psi_{div}}\)}&\scriptsize {\(0.99807\)} &\scriptsize \( 0.94750 \)
    &\scriptsize \(0.99790\) & \scriptsize \( 0.97250 \)& \scriptsize \(0.40437\) & \scriptsize \(0.10306\) \\ %
    \bottomrule
    \end{tabular}
\end{table*}

\begin{table*}%
    \caption{Analysis results of norm ratio, con-alpha ratio, div-alpha ratio, and con-div ratio on 5-shot task for \emph{mini}-ImageNet data with ResNet18 backbone. %
    }%
    \label{tab:miniImg_ResNet18_proposed_ratios_5s}
    \centering
    \begin{tabular}{M{2.65cm}|\Mc{1.25cm}\Mc{1.25cm}|\Mc{1.25cm}\Mc{1.25cm}|\Mc{1.75cm}|\Mc{1.75cm}}
    \toprule
    \scriptsize Measures & \multicolumn{2}{c}{\scriptsize ProtoNet\_S}& \multicolumn{2}{c}{\scriptsize ProtoNet\_{DR}}& \scriptsize Mann–Whitney U test& \scriptsize Fisher's exact test \\ 
    \scriptsize { }&\scriptsize Geometric mean &\scriptsize %
    Proportion & \scriptsize Geometric mean &\scriptsize %
    Proportion & & \\ \midrule
    \scriptsize {Norm ratio \(\phi\)}&\scriptsize {\(0.99925\)} &\scriptsize {\( \)} & \scriptsize \(0.99914\) & \scriptsize \( \) & \scriptsize \( 0.22914\) & \scriptsize \( \) \\  \hline
    \scriptsize {Con-alpha ratio \(\frac{\psi_{con}}{\hat{\alpha^*}}\)}&\scriptsize {\(0.99465\)} &\scriptsize \( 0.96250 \)
    &\scriptsize \(0.99554\) & \scriptsize \( 0.97500 \)& \scriptsize \( \mathbf{0.00810}\) & \scriptsize \( 0.41691\) \\ \hline
    \scriptsize {Div-alpha ratio \(\frac{\psi_{div}}{\hat{\alpha^*}}\)}&\scriptsize {\(1.00163\)} &\scriptsize \( 0.72750 \)
    &\scriptsize \(1.00195\) & \scriptsize \( 0.95000  \)& \scriptsize \(0.20071\) & \scriptsize \( \mathbf{<10^{-17}}\) \\ \hline
    \scriptsize {Con-div ratio \(\frac{\psi_{con}}{\psi_{div}}\)}&\scriptsize {\(0.99303\)} &\scriptsize \( 0.99000 \)
    &\scriptsize \(0.99360\) & \scriptsize \( 1.0000 \)& \scriptsize \(0.01124\) & \scriptsize \( 0.12406\) \\ %
    \bottomrule
    \end{tabular}
\end{table*}

\begin{figure*}
    \centering
    \begin{subfigure}[b]{0.24\textwidth}
        \includegraphics[width=1.0\linewidth,height=0.625\linewidth]{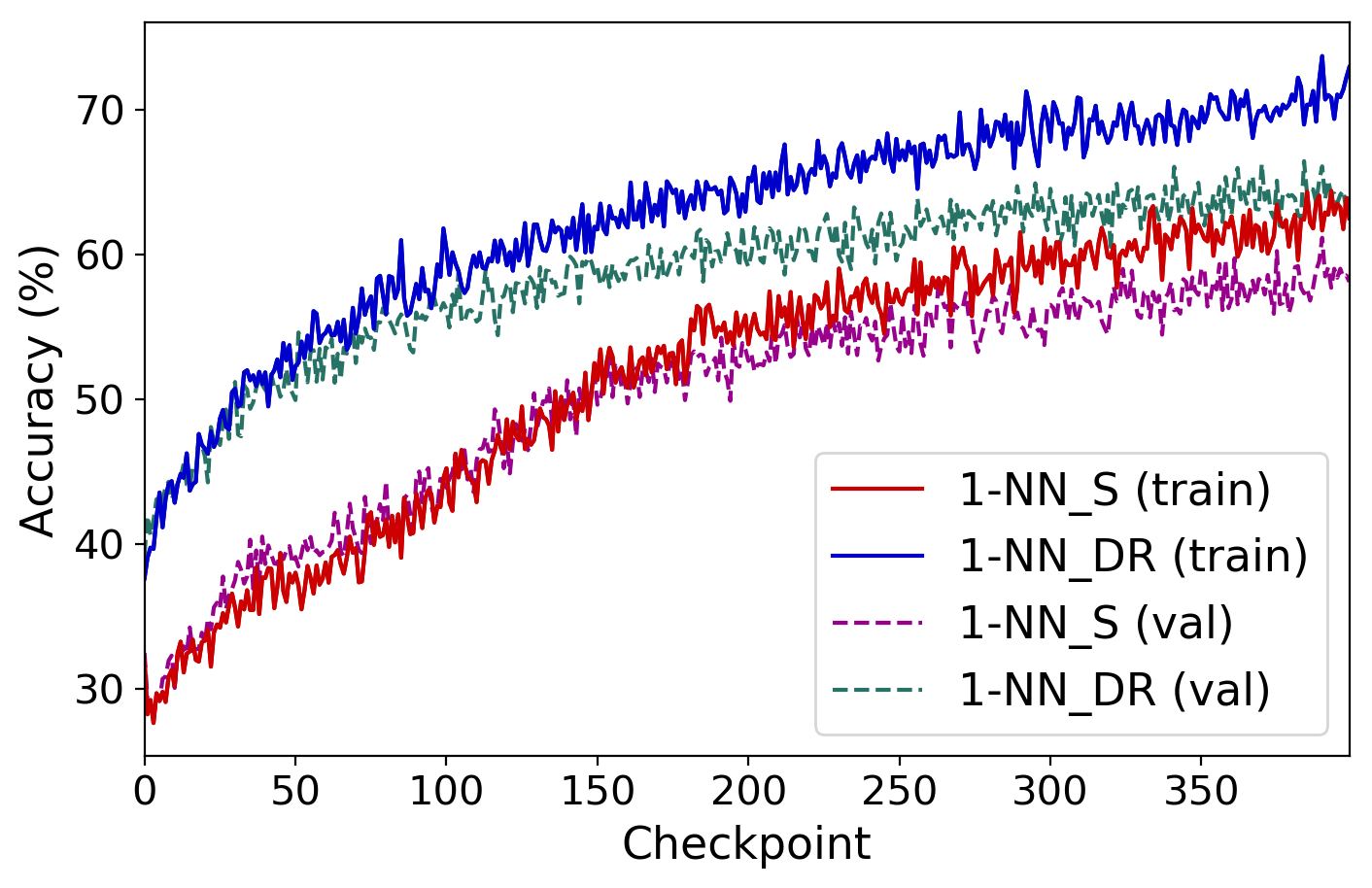} %
        \caption{Data: CUB, Backbone: Conv4}\label{fig:1nn_5-shot_training_a}
    \end{subfigure}
    \begin{subfigure}[b]{0.24\textwidth}
        \includegraphics[width=1.0\linewidth,height=0.625\linewidth]{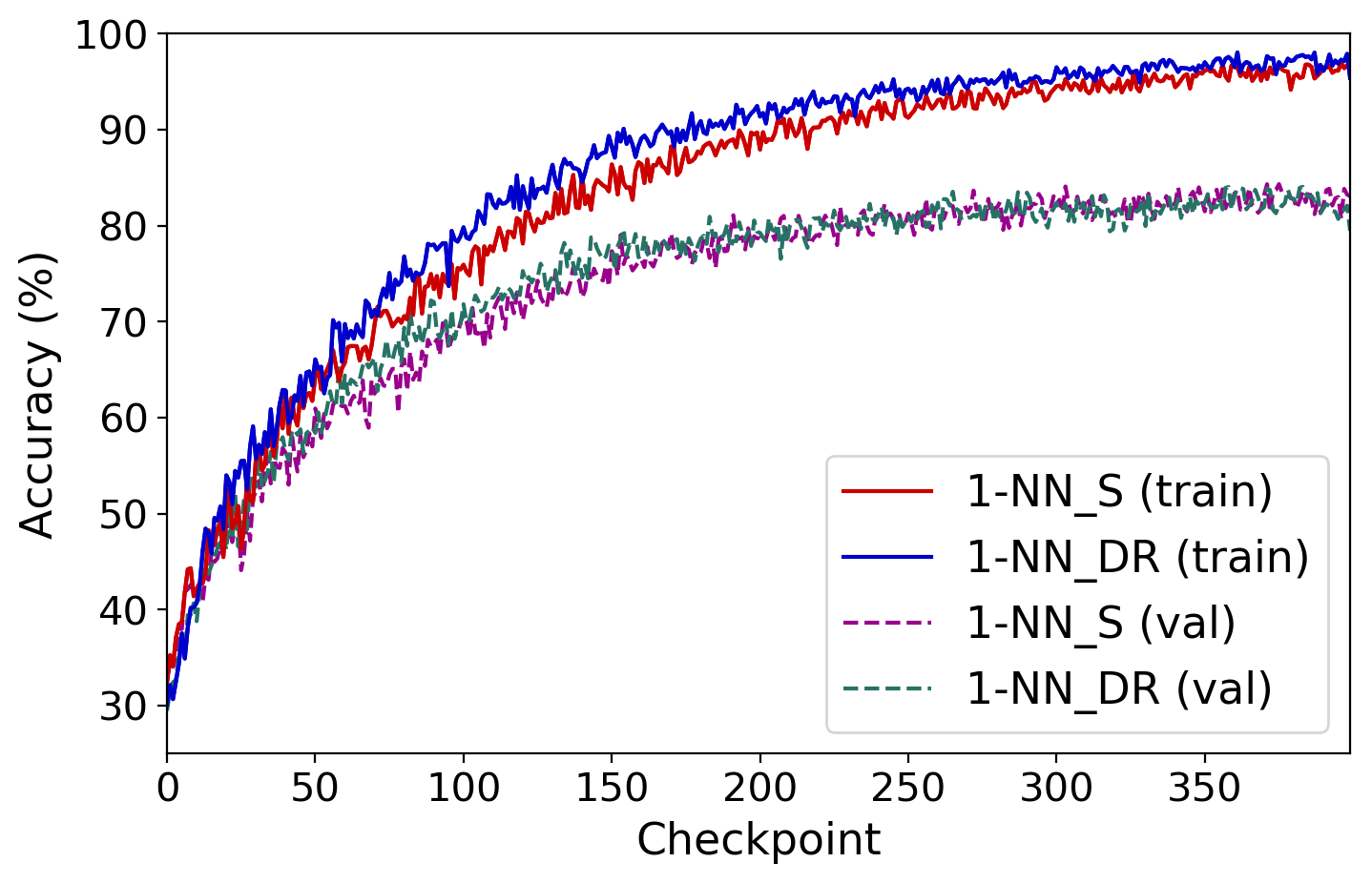}
        \caption{Data: CUB, Backbone: ResNet18}\label{fig:1nn_5-shot_training_b}
    \end{subfigure}
    \begin{subfigure}[b]{0.24\textwidth}
        \includegraphics[width=1.0\linewidth,height=0.625\linewidth]{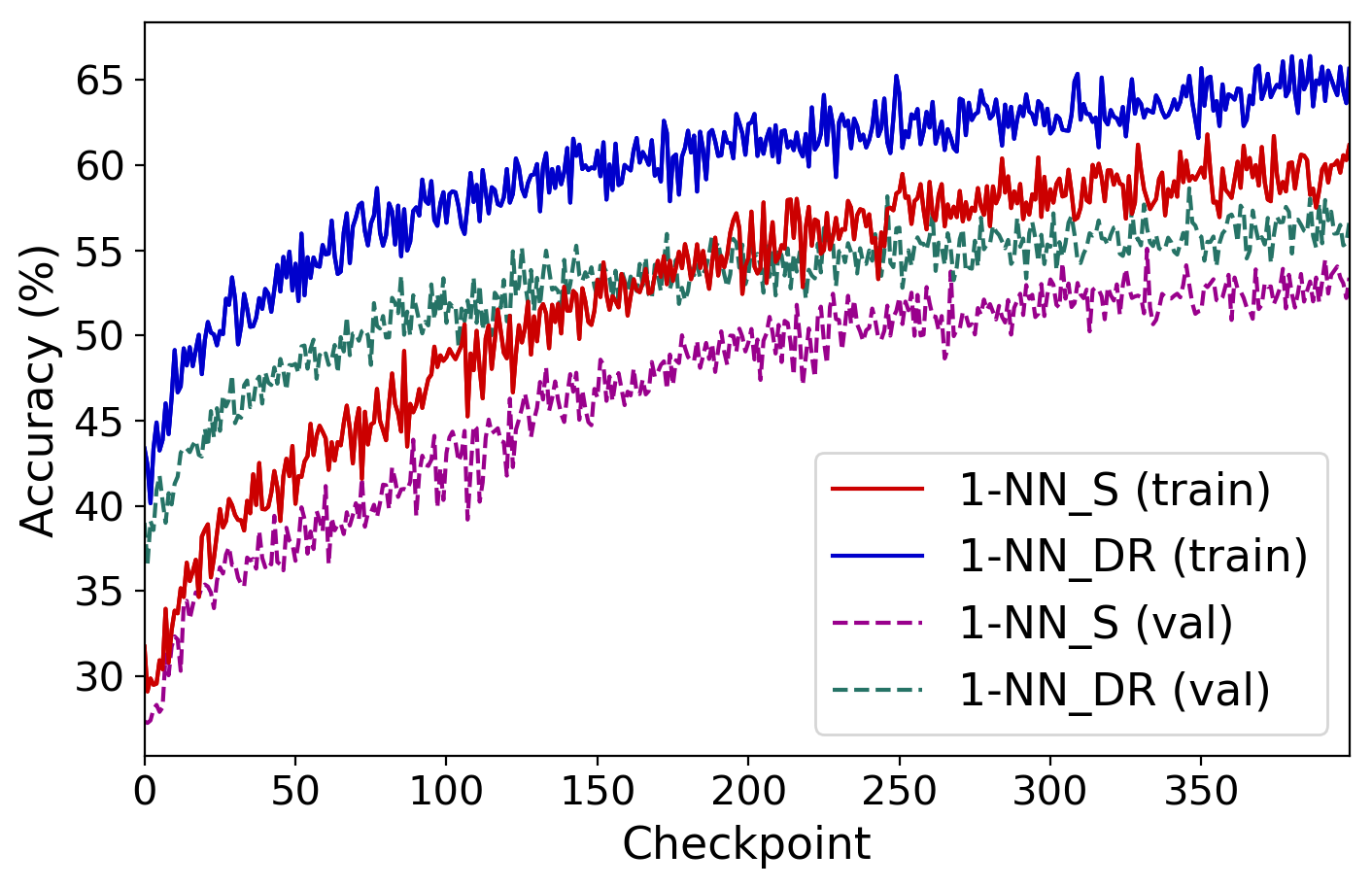} %
        \caption{Data: \emph{mini}-ImageNet, Backbone: Conv4}\label{fig:1nn_5-shot_training_c}
    \end{subfigure}
    \begin{subfigure}[b]{0.24\textwidth}
        \includegraphics[width=1.0\linewidth,height=0.625\linewidth]{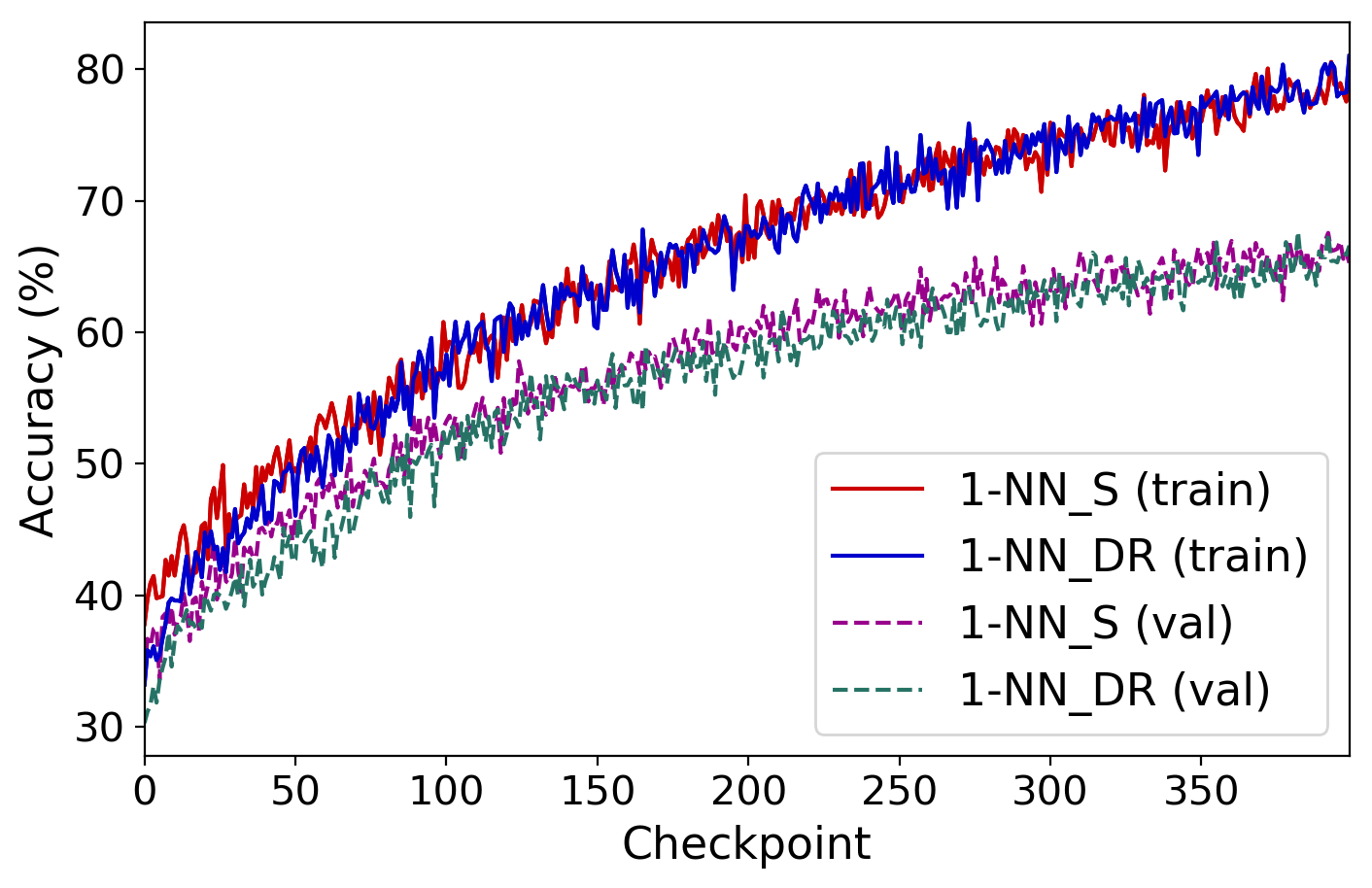}
        \caption{Data: \emph{mini}-ImageNet, Backbone: ResNet18}\label{fig:1nn_5-shot_training_d}
    \end{subfigure}
    \caption{Training and validation accuracy curves for two different backbones on 5-shot classification task (experiment with nearest neighbors based model). %
    }
    \label{fig:1nn_5-shot_training}
\end{figure*}

\newpage
\section{5-shot Learning Experiment with Nearest-Neighbor based Model}
\label{sec:1-NN_5s}

5-shot learning with the ResNet18 backbone (in Figures 3b %
and 3d) %
showed a slightly slower training speed when using DR formulation than using softmax-based formulation. To investigate if the reason is that %
a mean point is not an optimal point to represent a class in formulation (explained in Appendix \ref{sec:represent_mean_DR}), we try an additional experiment using nearest neighbors (1-NN) instead of the prototypical network \citep{snell2017prototypical}. %
In detail, in stead of using a prototype to define a distance to a class as in Equation (2), %
we use a distance defined as:
\begin{flalign}
d_{x';c}=\min\limits_{(x_i,y_i)\in S_c} d(f_{\theta}(x'),f_{\theta}(x_i)), \nonumber
\end{flalign}
where \(S_{c}\) is a set of support points with class \(c\). 
This distance represents a distance to the nearest support point on embedding space. Thus, the corresponding model becomes a differentiable nearest neighbor (1-NN) classifier. %
Here, we train differentiable 1-NN classifiers for 5-shot classification based on two formulations for each experiment: softmax-based (1-NN\_S) and distance-ratio-based (DR) formulation (1-NN\_{DR}).

Figure \ref{fig:1nn_5-shot_training} shows the training and validation accuracy curves when we use differentiable 1-NN classifiers. It shows that using DR formulation enables faster (Figure \ref{fig:1nn_5-shot_training_a}, \ref{fig:1nn_5-shot_training_b}, \ref{fig:1nn_5-shot_training_c}) or almost comparable (slightly slower) (Figure \ref{fig:1nn_5-shot_training_d}) training also for 5-shot tasks. %

\newpage
\section{An Example of a Normalized Embedding Space}\label{sec:normalized_ex}

To show the limitation of using cosine similarity in normalized embedding, we consider an example on a normalized embedding space \(\mathbb{S}^{2}\). \(r_1=(0,0,1)\), \(g_1=(0,1,0)\), and \(b_1=(1,0,0)\) are parameter vectors that represent red, green, and blue classes, respectively. In this example, we consider estimated probabilities \(\hat{p}(y=\textcolor{red}{red}|x')\) and the maximizing and minimizing positions using different models. (We used numerical optimization to get maximizing and minimizing positions of \(\hat{p}(y=\textcolor{red}{red}|x')\).)

\begin{figure}[H]
    \centering
    \begin{subfigure}[b]{0.24\textwidth}
        \includegraphics[width=1.0\linewidth,height=0.75\linewidth]{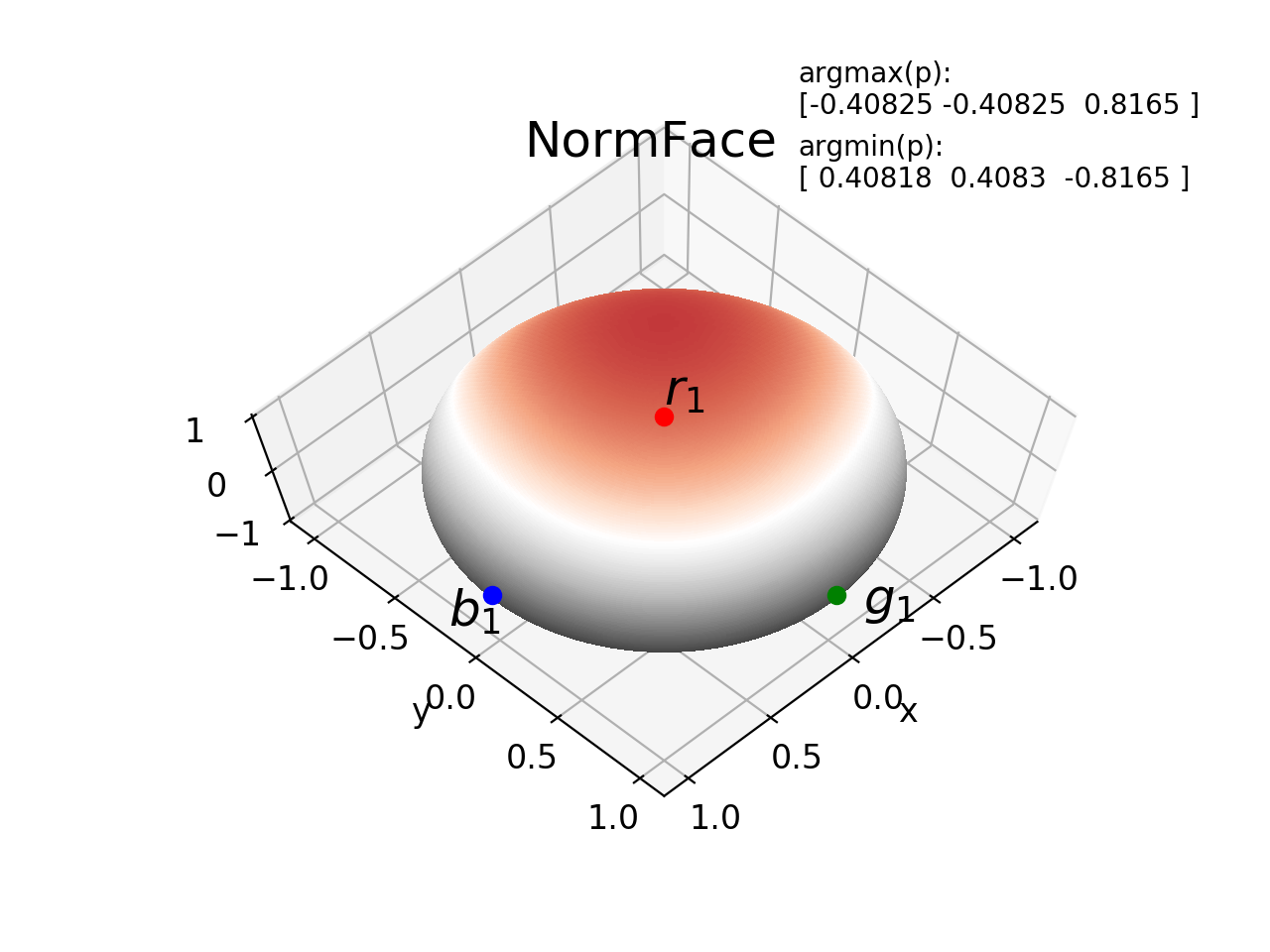}
        \caption{} %
    \end{subfigure}
    \begin{subfigure}[b]{0.24\textwidth}
        \includegraphics[width=1.0\linewidth,height=0.75\linewidth]{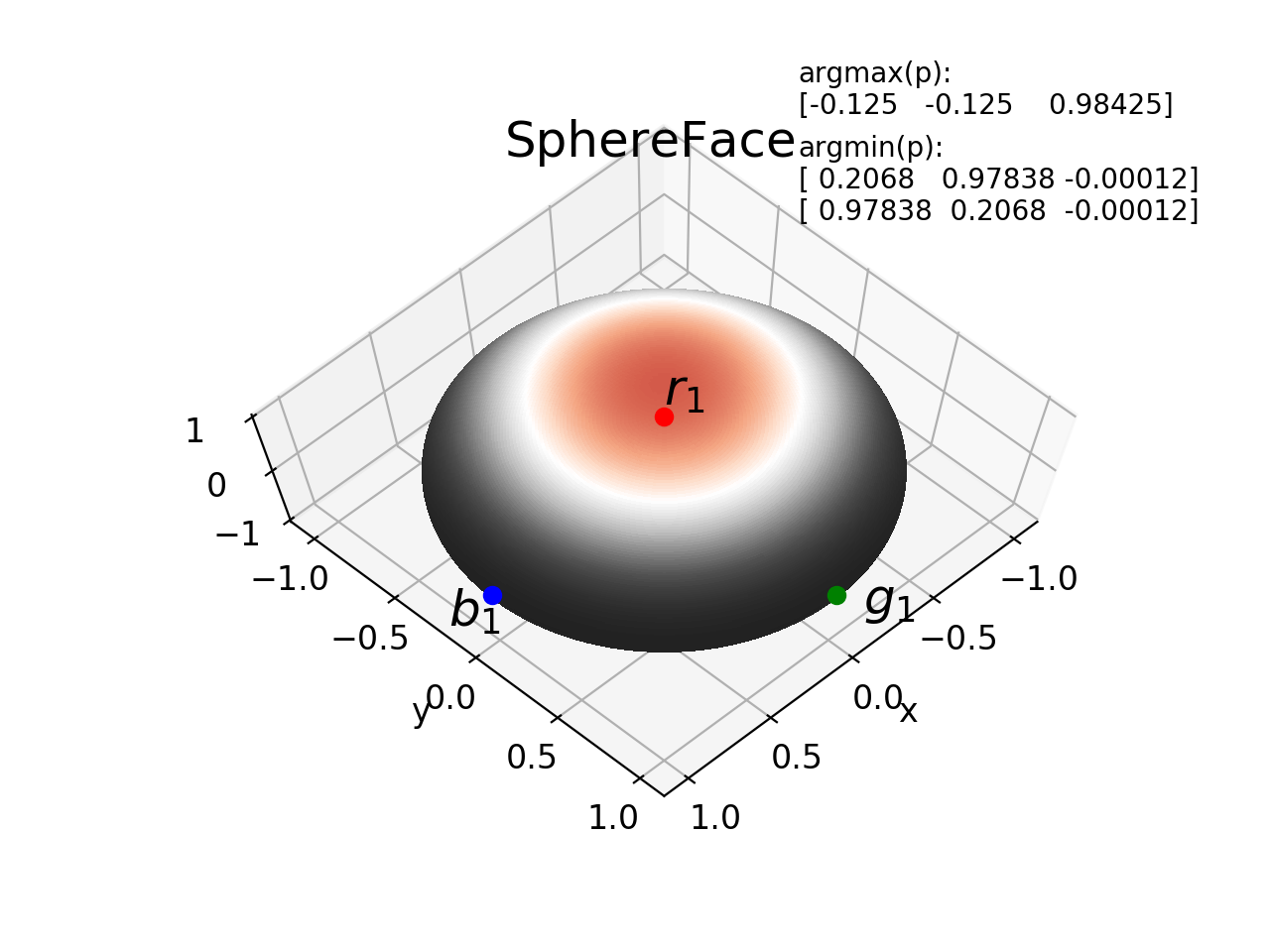}%
        \caption{}%
    \end{subfigure}
    \begin{subfigure}[b]{0.24\textwidth}
        \includegraphics[width=1.0\linewidth,height=0.75\linewidth]{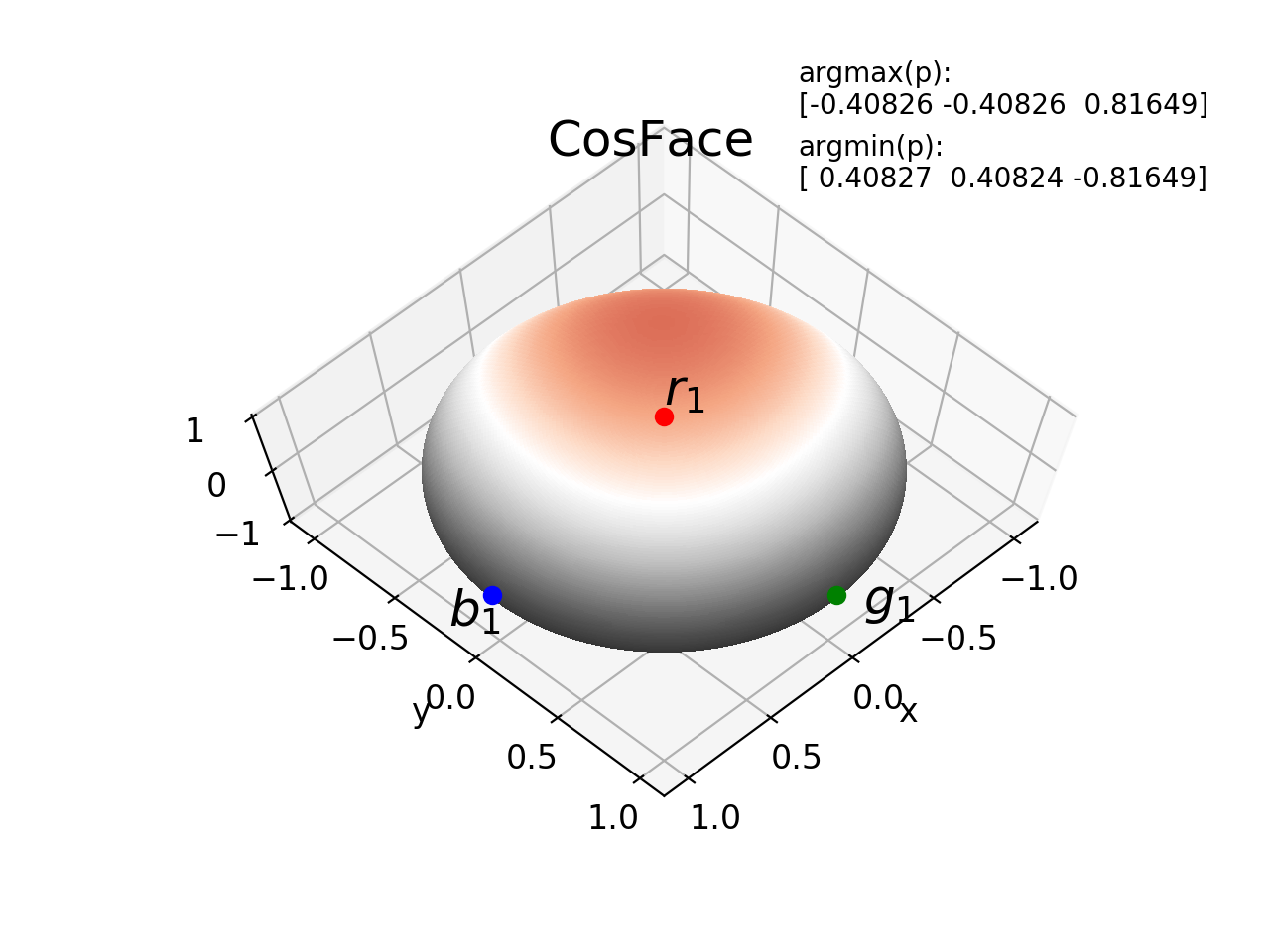}%
        \caption{}%
    \end{subfigure}
    \begin{subfigure}[b]{0.24\textwidth}
        \includegraphics[width=1.0\linewidth,height=0.75\linewidth]{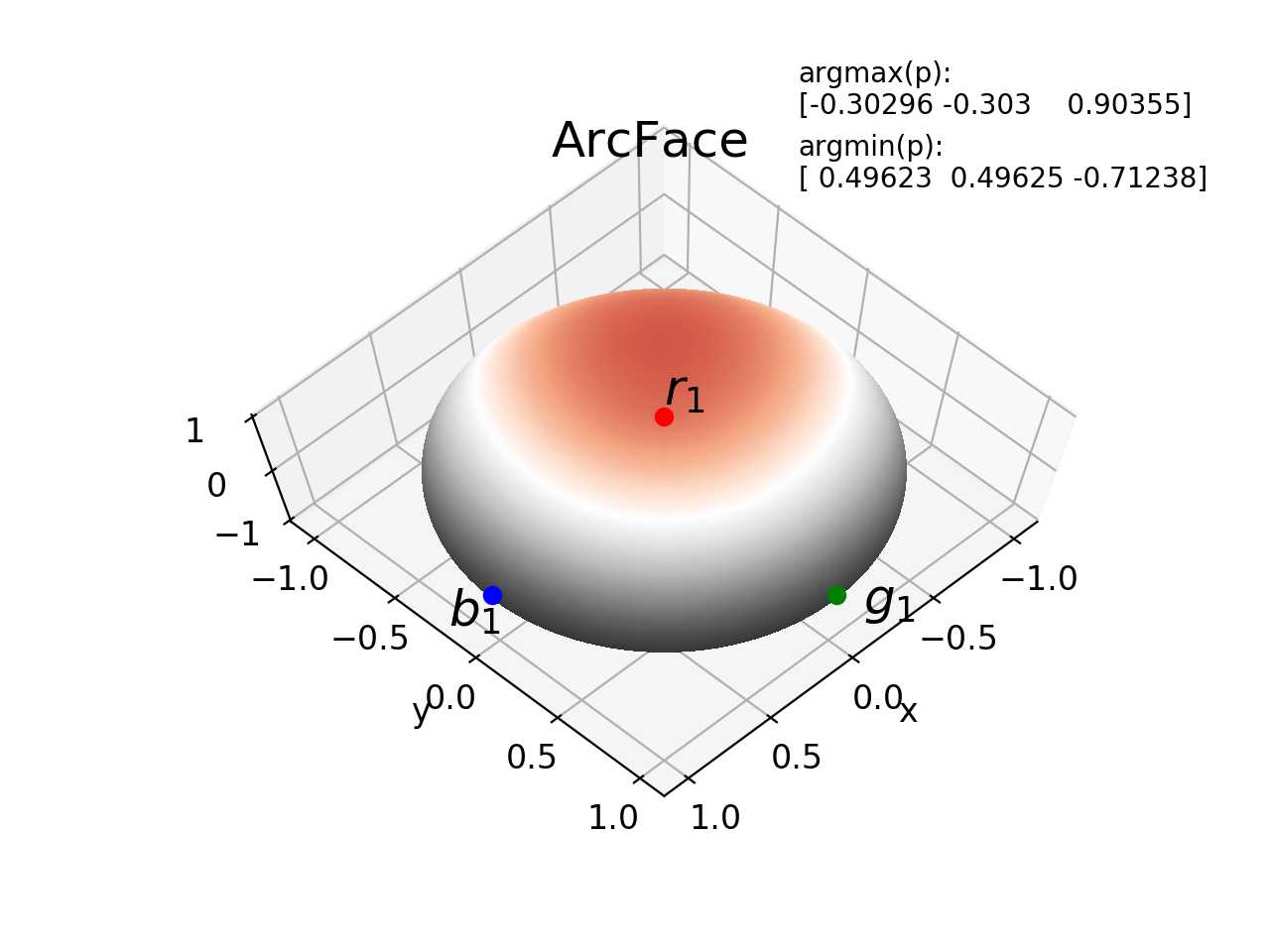}%
        \caption{}%
    \end{subfigure}
    \caption{Visualization of \(\hat{p}(y=\textcolor{red}{red}|x')\) for normalized %
    embedding. %
    Color is close to black when the value is small (close to zero). Color is close to dark red when the value is large (close to one). Models use cosine similarity. We used scaling parameter \(2.0\) for all models and margin \(2\) (with \(k=0\)) in SphereFace \citep{liu2017sphereface}.
    We used \(0.25\) as margins in CosFace \citep{wang2018cosface} and ArcFace \citep{deng2019arcface}.%
    }
    \label{fig:normalized_cos_sim}
\end{figure}

Figure \ref{fig:normalized_cos_sim} shows %
\(\hat{p}(y=\textcolor{red}{red}|x')\) and the maximizing and minimizing positions for models that use cosine similarity (NormFace \citep{wang2017normface}, SphereFace \citep{liu2017sphereface}, CosFace \citep{wang2018cosface}=AM-softmax \citep{wang2018additive}, ArcFace \citep{deng2019arcface}). %
All cosine similarity-based models take a different position than \(r_1\) as the position that maximizes \(\hat{p}(y=\textcolor{red}{red}|x')\). Also, these models take different positions than \(g_1\) and \(b_1\) as the positions that minimize \(\hat{p}(y=\textcolor{red}{red}|x')\).

\begin{figure}[H]
    \centering
    \begin{subfigure}[b]{0.325\textwidth}
        \includegraphics[width=1.0\linewidth,height=0.75\linewidth]{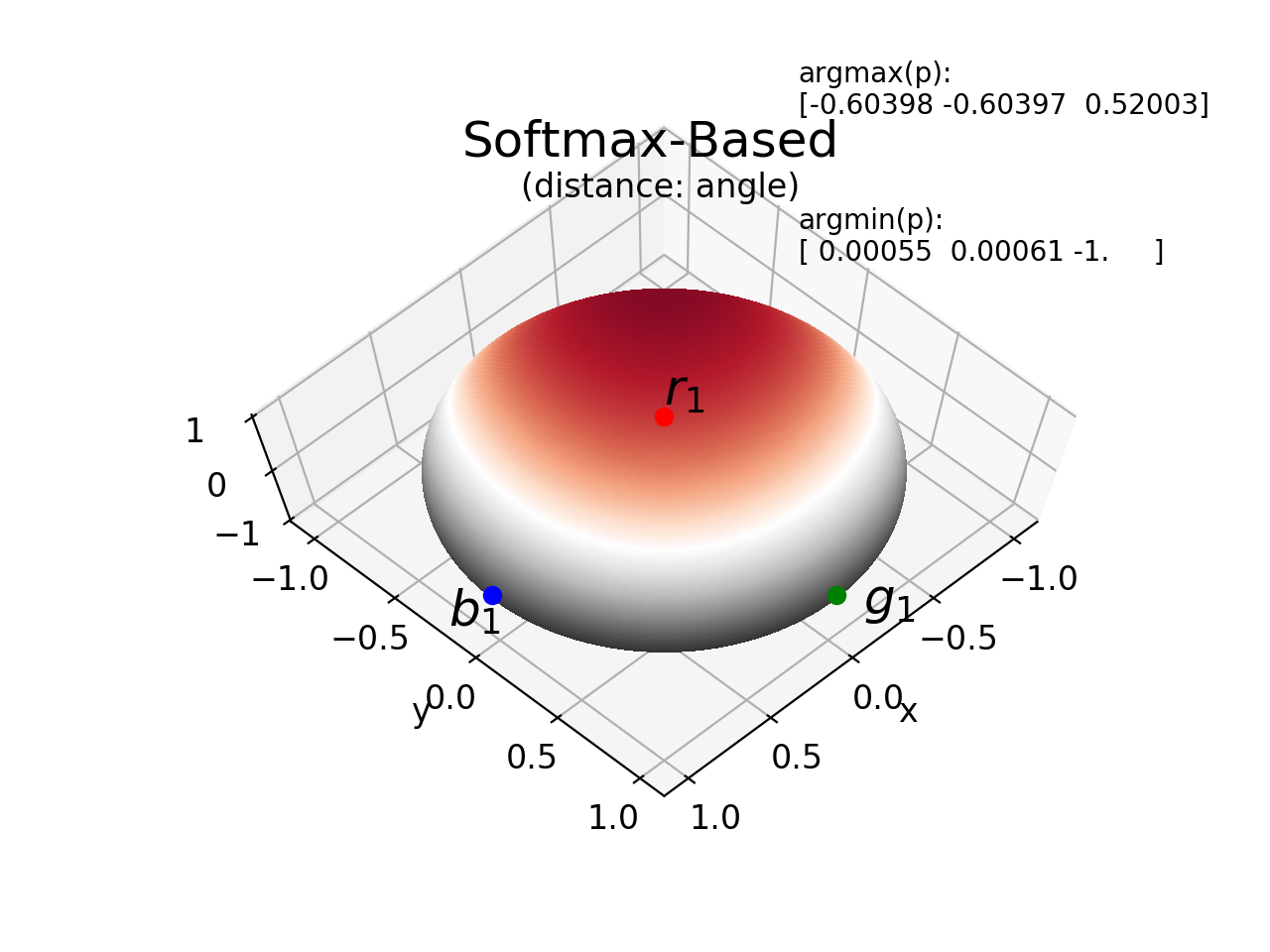}%
        \caption{}%
    \end{subfigure}
    \begin{subfigure}[b]{0.325\textwidth}
        \includegraphics[width=1.0\linewidth,height=0.75\linewidth]{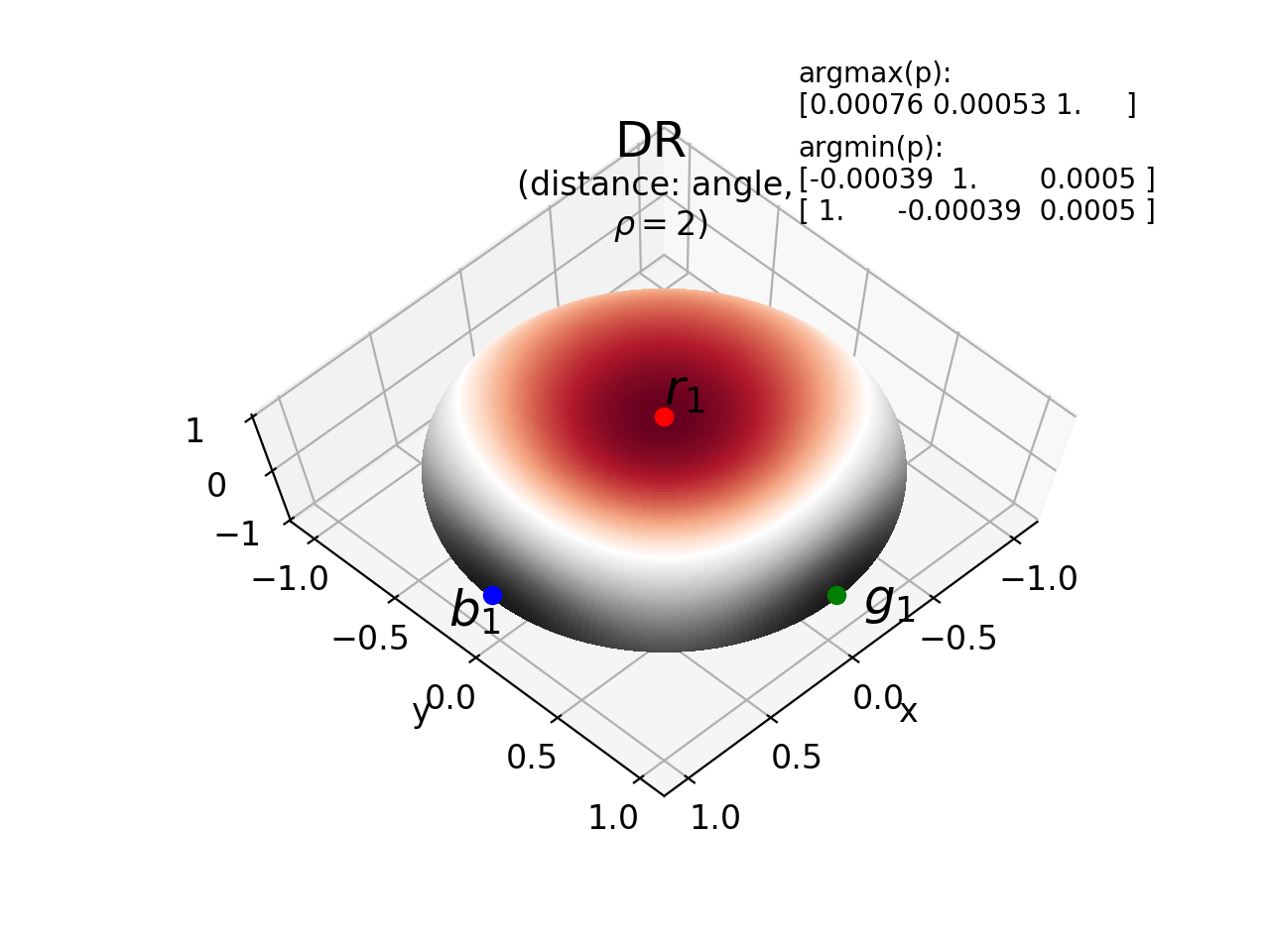}%
        \caption{}%
    \end{subfigure}
    \caption{Visualization of \(\hat{p}(y=\textcolor{red}{red}|x')\) for normalized %
    embedding. The same visualization was used as Figure \ref{fig:normalized_cos_sim}. Models use angular distance. We used \(\rho=2\) for DR formulation.}
    \label{fig:normalized_ang}
\end{figure}

Figure \ref{fig:normalized_ang} shows \(\hat{p}(y=\textcolor{red}{red}|x')\) and the maximizing and minimizing positions for models that use angular distance. In both models, we use angular distances with parameter vectors. %
By using angles (angular distances), we use the equations (4) and (5) to calculate \(\hat{p}(y=\textcolor{red}{red}|x')\).  The angle-based softmax-based model takes a different position than \(r_1\) as the position that maximizes \(\hat{p}(y=\textcolor{red}{red}|x')\). It takes \(-r_1\) as the position that minimizes \(\hat{p}(y=\textcolor{red}{red}|x')\). Angle-based DR formulation-based model takes \(r_1\) as the position that maximizes \(\hat{p}(y=\textcolor{red}{red}|x')\). It takes \(g_1\) and \(b_1\) as the positions that minimize \(\hat{p}(y=\textcolor{red}{red}|x')\). 
The example shows that using DR formulation %
helps data points to converge directly to the corresponding parameter vectors %
and data points can also diverge from parameter vectors for different classes.

\end{document}

%% file: math_commands.tex
\usepackage{amsmath,amsfonts,bm}

\def\eqref#1{equation~\ref{#1}}

\def\1{\bm{1}}

\DeclareMathAlphabet{\mathsfit}{\encodingdefault}{\sfdefault}{m}{sl}
\SetMathAlphabet{\mathsfit}{bold}{\encodingdefault}{\sfdefault}{bx}{n}

\DeclareMathOperator*{\argmin}{arg\,min}

\DeclareMathOperator{\Tr}{Tr}